\documentclass[11pt]{article}

\usepackage[preprint]{acl}

\usepackage{times}
\usepackage{latexsym}
\usepackage{amsmath}
\usepackage{booktabs}
\usepackage{multirow}
\usepackage{comment}
\usepackage{algorithm}
\usepackage{hyperref}
\usepackage{algpseudocode}
\usepackage{enumitem}
\usepackage[table]{xcolor}

\usepackage[T1]{fontenc}

\usepackage[utf8]{inputenc}

\usepackage{microtype}

\usepackage{inconsolata}

\usepackage{graphicx}
\usepackage{amssymb}
\title{DarkForest: Less Talk, Higher Accuracy for Multi-Agent LLMs}

\author{Yi Li$^{\Diamond}$,\quad Songtao Wei$^{\Diamond}$,\quad Dongming Jiang$^{\Diamond}$,\quad  Zhichun Guo$^{\clubsuit}$,\quad Qiannan Li$^{\spadesuit}$,\quad Bingzhe Li$^{\Diamond,}$\thanks{Corresponding author}\\
$^{\Diamond}$University of Texas at Dallas, $^{\clubsuit}$Independent Researcher, $^{\spadesuit}$University of California, Davis\\
\texttt{<Yi.Li3, Songtao.Wei, Dongming.Jiang, Bingzhe.Li>@utdallas.edu}\\
\texttt{zcguo.work@gmail.com},\quad \texttt{qnli@ucdavis.edu}}

\begin{document}
\maketitle
\begin{abstract}
Multi-agent LLM systems improve reasoning by combining outputs from multiple agents, but interaction-heavy methods can introduce error propagation and high communication overhead. When agents exchange raw responses or reasoning traces, incorrect intermediate reasoning may be adopted and amplified, leading to confident but wrong consensus; multi-round communication also increases token consumption, latency, and inference cost.
In this paper, we propose a controlled-communication coordination framework named \textbf{DarkForest}. DarkForest first keeps agents independent, so each agent produces an answer without seeing the others’ outputs. It then parses the raw responses into structured candidate records, groups semantically equivalent candidates into clusters, and estimates a calibrated belief distribution over these clusters using agent reliability, confidence, parse quality, support-pattern reliability, and independence corrections. A coordinator receives only policy-permitted evidence from this belief state with controlled communication. Experiments on six different reasoning benchmarks show that DarkForest achieves leading overall quality, improves the strongest baseline by up to 30.7\% on benchmark metrics, and reduces token consumption by up to 6.5$\times$ compared with communication-heavy baselines.

\end{abstract}

\section{Introduction}
Multi-agent large language model (LLM) systems~\cite{li2023camel,wu2024autogen,du2024improving} have become a prominent approach for improving test-time reasoning~\cite{yang2026towards}. Rather than relying on a single model instance~\cite{zhao2024expel}, these systems query multiple LLM-based agents and aggregate their outputs. Existing work has explored a wide range of interaction mechanisms~\cite{li2024survey}. Some systems use free-form conversation or role-playing to enable autonomous cooperation~\cite{li2023camel}, while others provide programming frameworks for building agents that communicate with one another~\cite{wu2024autogen}. Debate-based methods~\cite{du2024improving} ask agents to propose answers, exchange reasoning, and revise their conclusions over multiple rounds. Workflow-based systems, such as MetaGPT~\cite{hong2024metagpt}, assign agents specialized roles and coordinate their interaction through standard operating procedures. Collectively, these approaches show that agent interaction can improve task performance by eliciting diverse perspectives, enabling cross-checking, and supporting more structured reasoning.

\begin{figure}[t]
\centering
\includegraphics[width=\columnwidth]{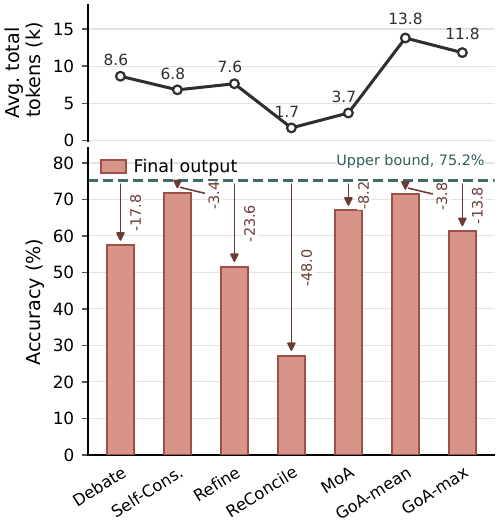}
\caption{
Token consumption and correct-candidate availability on the MATH dataset.
The top panel reports the total token consumption per sample for each baseline.
The bottom panel compares final accuracy (bars), with the candidate-availability upper bound (dashed line). This upper bound is defined as the percentage of examples for which at least one independently queried agent produces the correct answer. The gap indicates how often a coordination method fails to select a correct candidate that was already present among the initial agent outputs.}
\label{fig:motivation}
\vspace{-12pt}
\end{figure}

However, existing schemes face two major limitations. (1) Error propagation. Once an incorrect or misleading output is shared, later agents may adopt, refine, or amplify it, causing system to converge on a wrong answer with increasing confidence~\cite{huang2025survey,tyen2024llms}. In this case, agreement among agents no longer provides strong evidence of independent verification; instead, it may reflect imitation, persuasive influence, or contamination from earlier responses; (2) Communication overhead. Repeated message exchange across agents substantially increases token consumption, latency, and inference cost, making many multi-agent systems expensive to deploy at scale. These limitations raise a basic design question: \emph{what information should cross agent boundaries, and when does such disclosure improve, rather than contaminate, final decision-making?}\\
\textbf{Motivation: communication can lose correct evidence.} To test whether more communication reliably improves coordination, we examine whether final coordination preserves correct candidates that are already present in the initial agent outputs. For each example, we first check whether at least one independently queried agent predicts the correct answer. This measures whether the agent pool has already produced useful evidence before any cross-agent communication or aggregation. We then compare this correct-candidate availability with the final output accuracy of each coordination method. Figure~\ref{fig:motivation} shows various token consumptions and their final outputs remain far below the rate at which a correct candidate is initially available. This gap indicates that coordination can discard or overwrite useful independent evidence. Therefore, the central design question is not how to make agents communicate more, but how to control what information crosses agent boundaries.

To address these limitations, we draw inspiration from incomplete information game theory~\cite{aumann2002incomplete}. In settings marked by uncertainty, limited trust, costly communication, and high error cost, agents should not expose more information than is necessary for reliable coordination~\cite{vilares2011bayesian}. This principle reframes collaboration in multi-agent LLM systems. Agents need not share full reasoning traces in order to benefit from one another; instead, collaboration should be mediated by an explicit information policy~\cite{harsanyi1995games}. Such a policy specifies what information may cross agent boundaries, what evidence supports its disclosure, and when it is safe to incorporate that information into the final decision. Under this view, coordination is not equivalent to sharing more text; it is the controlled exposure of compact, policy-permitted, and verifiable evidence.


Following this principle, we propose \textbf{DarkForest}, which coordinates agent outputs through a structured belief interface. Each agent produces a response in isolation, which is parsed into a structured observation containing a canonical candidate, parse-validity status, confidence, and parse-quality metadata. DarkForest clusters equivalent candidates and constructs a calibrated belief distribution over candidate clusters, weighting support by agent reliability, support-pattern reliability, parse quality, confidence, and independence corrections. A disclosure policy then exposes only selected components of this belief state to the coordinator, such as parsed candidates, support patterns, confidence scores, posterior mass, and uncertainty indicators. The coordinator treats this exposed evidence as a prior rather than as proof, and a narrow deterministic guardrail intervenes only when the belief state strongly supports a candidate that conflicts with the coordinator output. Thus, DarkForest enables calibrated, policy-controlled coordination over compact evidence summaries. 

In summary, this paper makes the following contributions:
\begin{itemize}[topsep=0pt,itemsep=-1ex,partopsep=1ex,parsep=1ex, leftmargin=*]
    \item 
    We identify uncontrolled information exchange as a key challenge in multi-agent LLM reasoning, since unrestricted communication can amplify errors, compromise independent evidence, and increase inference cost.

    \item We propose \textbf{DarkForest}, a controlled communication coordination framework that agents generate candidate answers independently and coordination occurs through a calibrated belief state.

    \item A belief construction and disclosure mechanism are developed that combine parsing, candidate clustering, reliability calibration, confidence weighting, support-pattern modeling, and independence correction to expose only policy-permitted evidence to the coordinator.

    \item We evaluate DarkForest on six reasoning benchmarks spanning mathematics, code generation, general knowledge, scientific question answering, finance, and law with six different baselines. The code is open-sourced.\footnote{\url{https://github.com/PearLoveTana/DarkForest_Review}} 
\end{itemize}


\section{Background}
\textbf{Multi-agent LLM Collaboration.} Multi-agent LLM systems~\cite{li2023camel,wu2024autogen,du2024improving,chen2024reconcile,wang2022self,yun2026graph} query multiple language-model agents to solve the same task or different subparts of a task, then combine their outputs into a final answer. Existing systems instantiate this idea through different interaction patterns, including debate, round-table discussion, role-based workflows, graph-structured message passing, and aggregation over multiple sampled answers. These methods are effective because different agents may produce complementary reasoning paths, expose different failure modes, or specialize in different domains. However, they also differ in what information is exchanged: some methods expose full reasoning traces across agents, while others aggregate only final answers or candidate responses. We give a detailed comparison of these baselines in Section~\ref{sec:related} and Appendix~\ref{app: detailed_related}.\\
\textbf{How the Communication Works.} In multi-agent reasoning, communication can help agents correct mistakes, but it also changes the statistical meaning of agreement. If agents solve a problem independently, agreement among them provides evidence from multiple sources. If later agents observe earlier raw responses or reasoning traces, however, their outputs may no longer be independent: an incorrect but persuasive trace can be adopted, refined, or amplified by other agents. In this case, the system may converge to a confident but wrong consensus. Communication also has a direct systems cost. Multi-round exchange increases input length, output length, latency, and inference cost, especially when full reasoning traces are repeatedly copied into later prompts. Thus, a coordination mechanism should not treat communication as free or uniformly beneficial; it should decide what information is worth exposing.

\section{DarkForest Design}
This section presents the design of DarkForest. We first give the overall architecture (Section~\ref{sec:overall}) and the three design principles behind it: independent candidate generation, calibrated aggregation, and controlled communication. We then describe how raw agent responses are converted into structured observations (Section~\ref{sec：parsing}), how equivalent candidates are clustered (Section~\ref{sec: cluster}), and how DarkForest constructs a calibrated belief state over candidate clusters (Section~\ref{sec: belief}). Finally, we explain how the disclosure policy (Section~\ref{sec:disclosure}) exposes compact evidence to the coordinator and how the final decision rule combines the coordinator output with a narrow deterministic guardrail (Section~\ref{sec:coordinate}).

\subsection{Overall Architecture}\label{sec:overall}
\begin{figure*}[!t]
    \centering
    \includegraphics[width=6in]{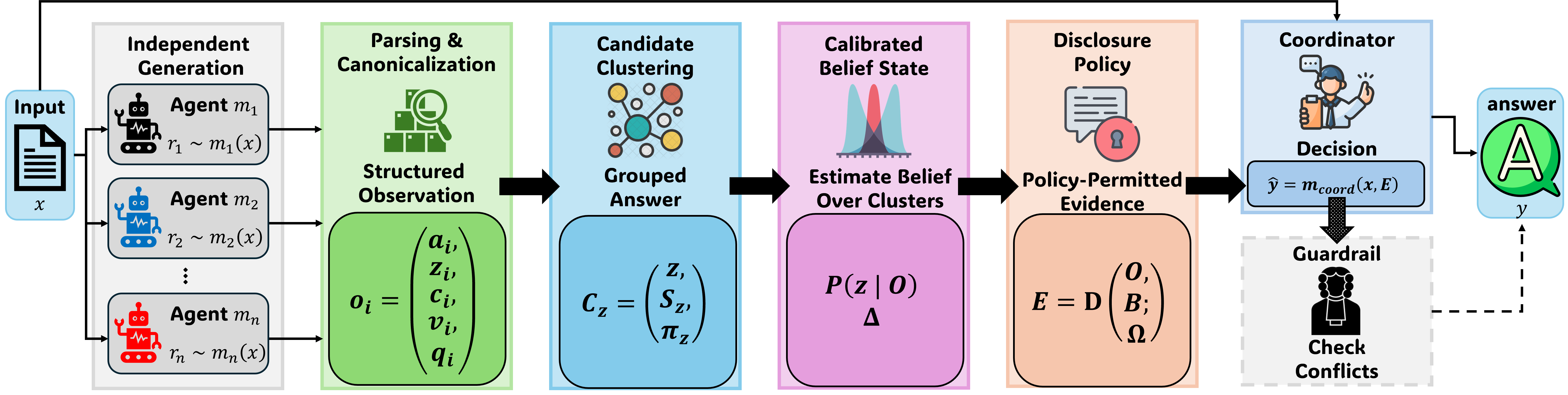}
    \caption{DarkForest pipeline. Agents independently generate candidate responses, which are parsed, canonicalized, and clustered into candidate hypotheses. DarkForest builds a calibrated belief state over these clusters and discloses only compact, policy-permitted evidence to the coordinator. The coordinator produces the final answer, with a deterministic guardrail that intervenes only when the belief state strongly supports a conflicting candidate.}
    \label{fig:overall}
\end{figure*}
DarkForest is a test-time coordination framework for aggregating the outputs of multiple LLM agents under uncertainty. The overall design philosophy is based on three principles, inspired by game theory with incomplete information: (1) Candidate generation should remain independent: agents do not observe one another’s outputs before producing their own; (2) Aggregation should be calibrated: agreement is weighted by historical reliability, parse quality, confidence, and agent dependence rather than treated as a uniform vote. (3) Communication should be controlled: the coordinator receives only policy-approved information, rather than unrestricted access to all raw traces.

As shown in Figure~\ref{fig:overall}, given an input $x$, a set of agents independently generates candidate outputs. DarkForest then parses these independent outputs into structured observations, clusters compatible candidates, constructs a calibrated belief state over the candidate clusters, and exposes a controlled summary to a final coordinator. Formally, let $\mathcal{M} = \{m_1,\cdots,m_n \}$ be the initial agent set. Each agent produces an output $r_i \sim m_i(x)$. DarkForest maps the raw outputs into structured observations, estimates a belief distribution over candidate clusters, and returns a final output $\hat{y} = F(x,r_1,\cdots,r_n)$. The function $F$ can be decomposed into parsing (Section~\ref{sec：parsing}), clustering (Section~\ref{sec: cluster}), belief construction (Section~\ref{sec: belief}), disclosure (Section~\ref{sec:disclosure}), coordination and deterministic correction (Section~\ref{sec:coordinate}).

\subsection{Parsing and Canonicalization}\label{sec：parsing}
\textbf{Independent Candidate.} Each agent receives the same input $x$ and independently generates a candidate response $r_i$. Agent do not receive other agents' candidates, confidence values, reasoning traces, or intermediate states. This preserves as much conditional independence as possible at inference time. We do not assume that all agents are fully independent in a statistical sense. Agents may share training data, architectures, or instruction-tuning procedures. The goal is narrower: the system avoids creating additional runtime dependence through iterative communication or mutual revision. This makes later agreement between agents more informative than it would be in the setting where agents directly influence one another.\\
\textbf{Parsing.} Raw model outputs are noisy objects. They may contain reasoning, invalid structure, formatting artifacts, or no usable final candidate. DarkForest therefore applies a task-specific parser $\mathcal{P}$ to each raw output $o_i = \mathcal{P}(r_i)$. Each parsed observation has the abstract form $o_i = (a_i,z_i,c_i,v_i,q_i)$, where $a_i$ is the parsed candidate, $z_i$ is a canonicalized candidate representation, $c_i\in [0,1]$ is the reported or imputed confidence, $v_i\in \{ 0,1 \}$ indicates parse validity, and $q_i$ stores parse-quality metadata such as malformedness or extraction method.\\
\textbf{Canonicalization} is domain-specific. Its purpose is to map semantically equivalent candidates to the same representation whenever possible. Thus coordination is performed over compact candidate records rather than over long free-form outputs. Invalid observations are excluded from candidate clustering: $V=\{ i:v_i=1 \}$. Malformed but parseable observations may remain usable, but their influence can be reduced during belief construction.

\subsection{Candidate Clustering}\label{sec: cluster}
DarkForest groups compatible parsed candidates into clusters. Let $z_i$ denote the canonical representation produced by agent $i$. For each distinct candidate $z$, define its support set as $S_{z} = \{ i\in V: z_i = z \}$. Each cluster is represented as $C_z = (z,S_z,\pi_z)$, where $\pi_z$ is the support pattern, i.e. the ordered identity of the agents supporting $z$. For example, $\pi_z$ records whether a candidate is supported by one agent, by a particular pair of agents, or by all agents. The cluster set is $\mathcal{C} = \{C_z : |S_z| > 0\}$. This step transforms heterogeneous free-form agent outputs into a finite set of competing candidate hypotheses.

\subsection{Calibrated Belief}\label{sec: belief}
DarkForest assigns each candidate cluster a calibrated evidence score. For a cluster $C_z$, the score is $s(z) = R_{\pi_z} \sum_{i\in S_z} \alpha_i \cdot \rho_i \cdot \delta_i \cdot \phi(c_i)$. Here, $\alpha_i$ is the calibrated reliability of agent $i$, $R_{\pi_z}$ is the calibrated reliability of support pattern $\pi_z$, $\rho_i$ is a parse-quality penalty, $\delta_i$ is an independence correction for correlated agents, and $\phi(c_i)$ maps confidence into a bounded evidence multiplier. The support-pattern term $R_{\pi_z}$ is applied at the cluster level because it estimates the reliability of the joint agreement pattern itself. For example, agreement between two complementary agents may provide stronger evidence than agreement between two highly correlated agents, even when their individual reliabilities are similar. Thus, DarkForest does not only ask how many agents support a candidate; it also asks which agents support it. The independence correction $\delta_i$ plays a different role from $R_{\pi_z}$. While $R_{\pi_z}$ calibrates the empirical reliability of the whole support pattern, $\delta_i$ discounts individual contributions when supporting agents are known to be correlated. This prevents correlated agents from adding evidence as if they were fully independent sources. The confidence multiplier is $\phi(c_i)=0.5+c_i$, so confidence scales an agent contribution between $0.5$ and $1.5$. This bounded affine form treats confidence as a weak modulation signal rather than a decisive vote. A low-confidence but valid candidate still contributes evidence, since it may be correct; a high-confidence answer can increase support, but cannot dominate the belief score without reliable agents or reliable support patterns. Thus confidence adjusts calibrated evidence without replacing calibration. The parse-quality penalty is
\[
\rho_i =
\begin{cases}
\lambda_{\mathrm{mal}}, & \text{output is malformed but parseable}\\
1, & \text{otherwise}
\end{cases}
\]
The independence correction satisfies $0 < \delta_i \leq 1$. It discounts agents known to be correlated with other supporting agents, preventing correlated agreement from being counted as fully independent evidence. Scores are normalized into a posterior-like belief distribution: $P(z \mid O) = \frac{\max(0,s(z))}
{\sum_{z' \in \mathcal{C}} \max(0,s(z'))}$, where $O=\{o_1,\ldots,o_n\}$. The top candidate is $z^\star = \arg\max_{z\in\mathcal{C}} P(z \mid O)$ and the posterior margin is
$\Delta = P(z^\star \mid O) - P(z^{(2)} \mid O)$, where $z^{(2)}$ is the second-highest posterior candidate. The belief state records $z^\star$, its posterior mass, the margin, the number of competing clusters, and whether the agents disagree.

DarkForest can use default parameters, but its intended use includes an offline calibration stage. Calibration runs the agents on examples with known outcomes and estimates the reliability terms used by the orchestrator. The language models themselves are not updated. For agent $i$, let $n_i^{\mathrm{valid}}$ be the number of valid parsed outputs and $n_i^{\mathrm{corr}}$ the number of correct ones. The agent reliability is estimated with Laplace smoothing: $\alpha_i
=
\frac{n_i^{\mathrm{corr}}+1}
{n_i^{\mathrm{valid}}+2}$. For support pattern $\pi$, let $n_\pi$ be the number of times the pattern appears and $n_\pi^{\mathrm{corr}}$ the number of times its supported cluster is correct. Its reliability is $R_\pi = \frac{n_\pi^{\mathrm{corr}}+1}
{n_\pi+2}$. If $n_\pi$ is below a minimum count, the pattern prior is not trusted and the system falls back to a default value. Missing-confidence behavior is calibrated as $c_{\mathrm{miss}} = \operatorname{clip}
\left(\frac{n_{\mathrm{miss}}^{\mathrm{corr}}}
{n_{\mathrm{miss}}},
c_{\min}, c_{\max}
\right)$. The malformed-output penalty is estimated by comparing malformed and well-formed accuracy: $\lambda_{\mathrm{mal}} = \operatorname{clip}
\left( \frac{\operatorname{Acc}_{\mathrm{mal}}}
{\operatorname{Acc}_{\mathrm{well}}},\lambda_{\min}, 1 \right)$. For a correlated pair of agents, an independence discount can be estimated from the incremental value of their agreement: $\gamma = \operatorname{clip}
\left( \frac{R_{ij}-\max(\alpha_i,\alpha_j)}
{1-\max(\alpha_i,\alpha_j)},\gamma_{\min}, 1 \right)$. Calibration therefore produces a frozen orchestrator parameter set. At evaluation time, DarkForest uses these parameters without modifying the agents or adapting on test examples.

DarkForest uses the belief distribution to identify unstable cases. A case is marked uncertain if the top posterior is low or if the leading candidate is insufficiently separated from the runner-up: $\operatorname{Uncertain}(O) = \mathbf{1} \left[P(z^\star \mid O) < \tau_u \lor \Delta < \tau_\Delta
\right]$. This uncertainty estimate does not by itself select the final output. Instead, it controls how the belief summary is presented to the coordinator. When the belief state is concentrated, the coordinator is encouraged to verify the leading cluster first. When the belief state is diffuse, the coordinator is encouraged to distrust simple agreement and audit candidates independently.

\subsection{Controlled Disclosure}\label{sec:disclosure}
The disclosure policy determines what information crosses from the initial agents to the coordinator. Let $E = D(O, B; \Omega)$ be the exposed evidence, where $B$ is the belief state and $\Omega$ is the disclosure policy. A restrictive policy may expose only candidate identifiers, canonical candidates, parse status, confidence, and a compact belief summary. A more permissive policy may expose reasoning summaries or truncated raw outputs. Full raw traces are not exposed unless explicitly allowed. DarkForest logs the disclosure cost for each instance:
$T_{\mathrm{cross}} = \operatorname{tokens}(E)$.
Thus communication is a measurable design variable rather than an implicit side effect of multi-agent prompting.

\subsection{Coordinator and Guardrail}\label{sec:coordinate}
The coordinator receives the original input and the exposed evidence: $r_{\mathrm{coord}} \sim m_{\mathrm{coord}}(x, E)$, and may follow the highest-posterior candidate, select a lower-posterior candidate, or synthesize a corrected output after checking the candidates against the input. The belief state is treated as a prior over candidates, not as proof. The coordinator output is parsed into a final candidate $\hat{z}_{\mathrm{coord}} = \mathcal{P}(r_{\mathrm{coord}})$. By default, DarkForest uses one coordinator call. Additional reflection or verifier calls are optional extensions rather than part of the base design.
\begin{algorithm}[!b]
\small
\caption{DarkForest Decision Procedure}
\label{alg:darkforest_decision}
\begin{algorithmic}[1]
\Require Input $x$; agents $\mathcal{M}=\{m_i\}_{i=1}^{n}$; parser $\mathcal{P}$; disclosure policy $\Omega$; calibrated parameters $\Theta$; coordinator $m_{\mathrm{coord}}$; thresholds $(k,\tau_p,\tau_m)$
\Ensure Final answer $\hat{y}$

\State \textbf{Independent generation:}
\For{$i=1,\ldots,n$}
    \State Generate response $r_i \sim m_i(x)$ without observing other agents.
    \State Parse $o_i=(a_i,z_i,c_i,v_i,q_i) \gets \mathcal{P}(r_i)$.
\EndFor

\State \textbf{Candidate clustering:}
\State $V \gets \{i : v_i=1\}$.
\State $\mathcal{C} \gets \{C_z=(z,S_z,\pi_z): S_z=\{i\in V:z_i=z\}, |S_z|>0\}$.

\State \textbf{Belief construction:}
\ForAll{$C_z \in \mathcal{C}$}
    \State $s(z) \gets R_{\pi_z}\sum_{i\in S_z}\alpha_i \rho_i \delta_i \phi(c_i)$ using $\Theta$.
\EndFor
\State $P(z\mid O) \gets \frac{\max(0,s(z))}{\sum_{z'\in\mathcal{C}}\max(0,s(z'))}$.
\State $z^\star \gets \arg\max_{z\in\mathcal{C}} P(z\mid O)$.
\State $\Delta \gets P(z^\star\mid O)-P(z^{(2)}\mid O)$, where $z^{(2)}$ is the runner-up candidate.
\State $B \gets (z^\star, P(z^\star\mid O), \Delta, \mathcal{C})$.

\State \textbf{Controlled disclosure and coordination:}
\State $E \gets D(O,B;\Omega)$.
\State $r_{\mathrm{coord}} \sim m_{\mathrm{coord}}(x,E)$.
\State $\hat{z}_{\mathrm{coord}} \gets \mathcal{P}(r_{\mathrm{coord}})$.

\State \textbf{Deterministic guardrail:}
\State $\mathrm{Trusted}(z^\star) \gets (|S_{z^\star}|\ge k)\wedge(P(z^\star\mid O)\ge\tau_p)\wedge(\Delta\ge\tau_m)$.
\If{$\mathrm{Trusted}(z^\star)$ and $\hat{z}_{\mathrm{coord}}\ne z^\star$}
    \State $\hat{y}\gets z^\star$.
\Else
    \State $\hat{y}\gets \hat{z}_{\mathrm{coord}}$.
\EndIf

\State \Return $\hat{y}$.
\end{algorithmic}
\end{algorithm}

The final decision rule combines the calibrated belief state with a single coordinator call and a narrow deterministic guardrail. Algorithm~\ref{alg:darkforest_decision} summarizes this procedure. The coordinator first receives only policy-permitted evidence and proposes a final answer. DarkForest then applies deterministic correction only when the belief state provides strong support for a conflicting candidate. This separation keeps language-model reasoning and deterministic evidence checks explicit: the coordinator can inspect compact evidence against the input, while the guardrail prevents strongly supported candidates from being discarded by a single coordinator call.\\
\textbf{Guardrail.} DarkForest applies a final deterministic guardrail, which intervenes only when the belief state strongly supports a candidate that conflicts with the coordinator output. A top cluster $z^\star$ is trusted if $(|S_{z^\star}| \geq k) \wedge (P(z^\star \mid O) \geq \tau_p) \wedge (\Delta \geq \tau_m)$. If these conditions hold, the guardrail may replace the coordinator output:
$$\hat{y} =
\begin{cases}
z^\star, & \text{if } z^\star \text{ is trusted and } \hat{z}_{\mathrm{coord}}\neq z^\star\\
\hat{z}_{\mathrm{coord}}, & \text{otherwise}
\end{cases}
$$
The guardrail is intentionally narrow. It introduces no additional model calls and uses only information already available in the belief state.

\section{Evaluation}
This section evaluates DarkForest across six reasoning benchmarks. We first describe the experimental setup (Section~\ref{sec:setup}). We then compare DarkForest with representative multi-agent baselines in terms of task quality and average token consumption (Section~\ref{sec:overall_comp}). To keep the main text focused, we include the key coordinator and guardrail ablation in Section~\ref{sec:coordinator_guardrail} and defer additional component-level ablations to Appendix~\ref{app:ablation}.

\subsection{Experimental Setup}\label{sec:setup}
We evaluate DarkForest on six benchmarks covering both general and professional domain reasoning: MATH~\cite{lightman2024let} for mathematical problem solving, HumanEval~\cite{chen2021evaluating} for code generation,  MMLU-Pro~\cite{wang2024mmlu} for broad multi-domain reasoning, GPQA~\cite{rein2023gpqa} for graduate-level scientific question answering, FinQA~\cite{chen2021finqa} for financial reasoning, and LegalBench~\cite{guha2023legalbench} for legal reasoning. We report the default metric for each benchmark: exact match for MATH and LegalBench, Pass@1 for HumanEval, accuracy for MMLU-Pro and GPQA, and both execution accuracy and program accuracy for FinQA. For each benchmark, DarkForest coordinates three independently queried agents. General-domain tasks use Qwen2.5-7B-Instruct~\cite{Yang2024Qwen25TR}, Qwen2.5-Coder-7B-Instruct~\cite{hui2024qwen2}, and Mathstral-7B-v0.1~\cite{mistral_mathstral_2024}; professional-domain tasks additionally use domain-specialized agents, including finance-Llama3-8B~\cite{xie2023pixiu} and Saul-7B-Instruct-v1~\cite{colombo2024saullm}. All models are served with vLLM~\cite{kwon2023efficient} under the same benchmark-specific decoding settings whenever the method permits. We compare against six state-of-art multi-agent baselines, including Debate~\cite{du2024improving}, Self-Consistency~\cite{wang2022self}, Refine~\cite{madaan2023self}, ReConcile~\cite{chen2024reconcile}, Mixture-of-Agent~\cite{wang2025mixture}, and Graph-of-Agent~\cite{yun2026graph}. Besides task quality, we report average input and output token consumption per sample to measure end-to-end cost. Full setup details are given in Appendix~\ref{app:detailed_setup}.

\subsection{Overall Comparison}\label{sec:overall_comp}
\begin{table*}[t]
\centering
\caption{Benchmark quality evaluation across general and professional domains. General-domain benchmarks include MATH, HumanEval, MMLU-Pro, and GPQA, evaluated with Qwen2.5-7B-Instruct, Qwen2.5-Coder-7B-Instruct, and Mathstral-7B-v0.1. Professional-domain benchmarks include FinQA and LegalBench, evaluated with Qwen2.5-7B-Instruct, finance-Llama3-8B, and Saul-7B-Instruct-v1. \textbf{\emph{The best result}} is marked in italic bold, and \underline{the second-best result} is marked with underline only.}
\label{tab:overall_comparison}
\resizebox{\textwidth}{!}{%
\begin{tabular}{lcccc|cc}
\toprule[1.5pt]
\multirow{2}{*}{}     
& \multicolumn{4}{c|}{\begin{tabular}[c]{@{}c@{}}General Domain\end{tabular}} 
& \multicolumn{2}{c}{\begin{tabular}[c]{@{}c@{}}Professional Domain \end{tabular}} \\ 
\cmidrule(lr){2-7} 
\begin{tabular}[c]{@{}l@{}}Benchmark\\ Default Quality Metric ($\mathbf{\%}$)\end{tabular}
& \begin{tabular}[c]{@{}c@{}}MATH\\ Exact Match\end{tabular}
& \begin{tabular}[c]{@{}c@{}}HumanEval\\ Pass@1\end{tabular}
& \begin{tabular}[c]{@{}c@{}}MMLU-Pro\\ Accuracy\end{tabular}
& \begin{tabular}[c]{@{}c@{}}GPQA\\ Accuracy\end{tabular}
& \begin{tabular}[c]{@{}c@{}}FinQA\\ Execution Accuracy $\|$ Program Accuracy\end{tabular}
& \begin{tabular}[c]{@{}c@{}}LegalBench\\ Exact Match\end{tabular} \\ 
\midrule[1pt]
Debate                & 57.40 & 64.00 & \underline{55.86} & 35.86 & 3.67 $\|$ 1.33  & 57.60 \\
Self-Consistency      & \underline{71.80} & 68.00 & 53.00 & \textbf{\emph{40.40}} & 6.33 $\|$ 3.33     & 65.00 \\
Refine                & 51.60 & 70.00 & 52.62 & 34.34 & 8.67 $\|$ 3.33 & 60.80 \\
ReConcile             & 27.20 & 16.00 & 45.71 & 34.34 & 4.33 $\|$ 1.33  & \textbf{\emph{69.00}} \\
Mixture-of-Agent      & 67.00 & 84.00 & 53.38     & 35.86 & 8.33 $\|$ 4.67 & 53.80 \\
Graph-of-Agent (Mean) & 71.40  & 80.00 & 40.71 & 34.85 & \textbf{\emph{16.00}} $\|$ \underline{8.67}     & 46.40 \\
Graph-of-Agent (Max)  & 61.40  & \textbf{\emph{86.00}} & 41.86 & 30.30 & 11.33 $\|$ 7.00     & 32.40 \\
\midrule[1pt] 
DarkForest (Ours)     & \textbf{\emph{76.80}}  & \underline{84.00} & \textbf{\emph{58.38}} & \underline{39.90} & \underline{15.67} $\|$ \textbf{\emph{11.33}}     & \underline{68.00} \\
\bottomrule[1.5pt]
\end{tabular}%
}
\end{table*}
\begin{figure*}
    \centering
    \includegraphics[width=\textwidth]{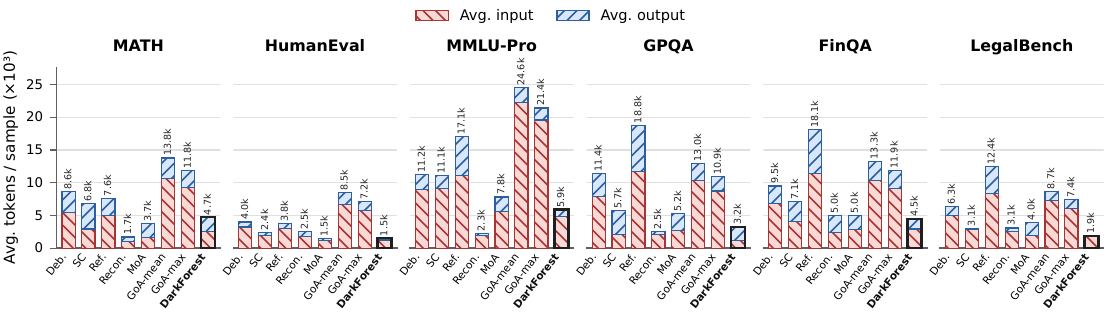}
    \caption{Average token consumption per sample across six benchmarks. 
Each stacked bar decomposes total token usage into input tokens and output tokens. Numbers above bars indicate total token usage in thousands.}
    \label{fig:token_consumption}
    \vspace{-12pt}
\end{figure*}

Table~\ref{tab:overall_comparison} reports the main task quality results across six benchmarks, covering mathematical reasoning, code generation, general knowledge reasoning, scientific question answering, financial reasoning, and legal reasoning. Figure~\ref{fig:token_consumption} reports the corresponding average token consumption per sample. Overall, DarkForest achieves the strongest or near-strongest quality across the benchmark suite while using substantially fewer tokens than communication-heavy multi-agent baselines.

DarkForest improves over the second strongest baseline by 2.52--5.00 absolute points, corresponding to a 4.5\%--30.7\% relative improvement. When it ranks second, the gap to the best result is consistently small, only 1.2\%--2.3\% relative to the best. On MATH, DarkForest improves exact match to 76.80\%, outperforming Self-Consistency by 5.00 points and Graph-of-Agent (Mean) by 5.40 points. On MMLU-Pro, DarkForest reaches 58.38\%, improving over Debate by 2.52 points and over all aggregation-based baselines by larger margins. On FinQA, DarkForest achieves the best program accuracy, 11.33\%, and the second-best execution accuracy, 15.67\%, only 0.33 points below Graph-of-Agent (Mean), and on LegalBench, DarkForest reaches 68.00\%, within 1.00 point of ReConcile and above the remaining baselines. These results show that DarkForest is not specialized to one task type: the same selective-communication design remains competitive across mathematical, general, scientific, financial, and legal reasoning settings.

Several data points highlight why the design matters. First, on MATH, DarkForest outperforms both simple independent aggregation and graph-based communication. This suggests that the gain does not come merely from querying multiple agents, but from calibrating which agents support which candidate and treating agreement as structured evidence rather than as a uniform vote. Second, HumanEval is the only one where DarkForest is not among the top standalone methods: it matches Mixture-of-Agent at 84.00\% and trails Graph-of-Agent (Max) by 2.00 points. This is consistent with the nature of code generation, where preserving richer candidate programs or implementation details can be useful. DarkForest still remains competitive while using only 1.5k tokens per sample, compared with 8.5k and 7.2k for the two Graph-of-Agent variants. Third, in professional domains, DarkForest performs well without exposing long raw responses: it obtains the best FinQA result and near-best LegalBench exact match. This supports the central premise of DarkForest: in many reasoning tasks, agents need not exchange full reasoning traces for coordination to be effective; compact, calibrated evidence is often sufficient.

Figure~\ref{fig:token_consumption} further shows that DarkForest improves the task quality and cost. Compared with Graph-of-Agent (Mean), DarkForest reduces average token usage from 13.8k to 4.7k on MATH, from 24.6k to 5.9k on MMLU-Pro, from 13.0k to 3.2k on GPQA, and from 8.7k to 1.9k on LegalBench. These reductions follow directly from DarkForest's disclosure policy: initial agents generate candidates independently, and the coordinator receives a compact belief summary instead of full raw traces. Thus, the overall comparison supports the main hypothesis of DarkForest: more inter-agent communication is not necessary for strong multi-agent reasoning; what matters is preserving independent evidence, calibrating agreement, and disclosing only the information needed for the final decision.

\subsection{Coordinator and Guardrail}\label{sec:coordinator_guardrail}
\begin{table}[t]
\centering
\caption{Decision mechanism ablation. w/o Coordinator directly returns the top-belief candidate without a coordinator call, and w/o Guardrail keeps the coordinator but removes the deterministic post-processing step.}
\label{tab:ablation_decision}
\resizebox{\linewidth}{!}{%
\begin{tabular}{lccc}
\toprule[1.5pt]
\multirow{2}{*}{Dataset}
& \multicolumn{3}{c}{Quality} \\
\cmidrule(lr){2-4}
& \begin{tabular}[c]{@{}c@{}}DarkForest\\w/o Coordinator\end{tabular}
& \begin{tabular}[c]{@{}c@{}}DarkForest\\w/o Guardrail\end{tabular}
& \begin{tabular}[c]{@{}c@{}}Full\\DarkForest\end{tabular} \\
\midrule[1pt]
MATH       & 76.80 & 75.40 & 76.80 \\
LegalBench & 67.20 & 65.60 & 68.00 \\
\bottomrule[1.5pt]
\end{tabular}%
}
\end{table}
\begin{figure}[t]
\centering
\includegraphics[width=\columnwidth]{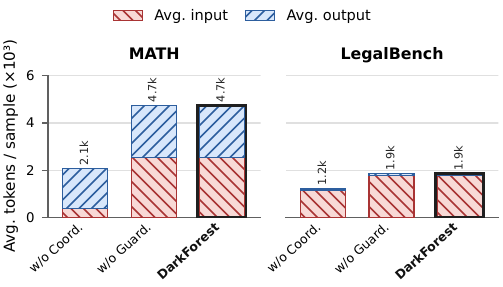}
\caption{Average token consumption for the decision mechanism ablation.}
\label{fig:ablation_decision_tokens}
\end{figure}
We now isolate the two final-decision components of DarkForest: the coordinator and the deterministic guardrail. Table~\ref{tab:ablation_decision} compares Full DarkForest with two variants. The first removes the coordinator and directly returns the top-belief candidate. The second keeps the coordinator but removes the deterministic guardrail. The results show that the coordinator and guardrail play complementary roles. Removing the coordinator has no effect on MATH, where the calibrated belief state is already sufficiently decisive, but reduces LegalBench accuracy from 68.00\% to 67.20\%. This suggests that the coordinator is most useful when candidate labels are heterogeneous or when the task benefits from checking compact evidence against the original input. In contrast, removing the guardrail hurts both benchmarks: MATH drops from 76.80\% to 75.40\%, and LegalBench drops from 68.00\% to 65.60\%. The guardrail therefore protects the system from cases where the coordinator output conflicts with a strongly supported belief cluster. Figure~\ref{fig:ablation_decision_tokens} reports the corresponding token cost. The variant without a coordinator is cheaper because it avoids the final coordinator call, using 2.1k tokens on MATH and 1.2k on LegalBench. Full DarkForest uses the same token budget as the no-guardrail variant because the guardrail is deterministic post-processing and introduces no model call. This confirms that the guardrail improves quality without increasing token consumption.

\section{Related Work}
\label{sec:related}

Multi-agent LLM coordination methods differ primarily in what information crosses agent boundaries. Debate-based methods~\cite{du2024improving,chen2024reconcile} expose full reasoning traces and revise answers over multiple rounds, which makes them vulnerable to error propagation when a persuasive but incorrect trace influences subsequent turns. Aggregation-based methods~\cite{wang2022self} combine candidate outputs but largely treat agreement as an unweighted vote, without explicitly modeling per-agent reliability, parse quality, or inter-agent dependence. Refinement and efficiency approaches~\cite{madaan2023self,wang2025agentdropout} target output quality or communication cost separately, without treating disclosure itself as a controlled design variable. We give an expanded treatment of each baseline in Appendix~\ref{app: detailed_related}.

\section{Conclusion}
We presented DarkForest, a controlled communication framework for multi-agent LLM. DarkForest keeps candidate generation independent, converts raw agent outputs into structured observations, builds a calibrated belief state over candidate clusters, and exposes only policy-permitted evidence to a coordinator. A deterministic guardrail further prevents strongly supported candidates from being discarded by the coordinator. Experiments across six benchmarks show that DarkForest achieves strong or near-strong task quality while using substantially fewer tokens than communication-heavy baselines. The ablations further show that DarkForest's gains come from the combined effect of its core components. Overall, our results suggest that the key question in multi-agent LLM reasoning is not how to make agents communicate more, but how to preserve independent evidence and control what information crosses agent boundaries.

\newpage
\bibliography{custom}

@article{li2023camel,
  title={Camel: Communicative agents for" mind" exploration of large language model society},
  author={Li, Guohao and Hammoud, Hasan and Itani, Hani and Khizbullin, Dmitrii and Ghanem, Bernard},
  journal={Advances in neural information processing systems},
  volume={36},
  pages={51991--52008},
  year={2023}
}

@inproceedings{wu2024autogen,
  title={Autogen: Enabling next-gen LLM applications via multi-agent conversations},
  author={Wu, Qingyun and Bansal, Gagan and Zhang, Jieyu and Wu, Yiran and Li, Beibin and Zhu, Erkang and Jiang, Li and Zhang, Xiaoyun and Zhang, Shaokun and Liu, Jiale and others},
  booktitle={First conference on language modeling},
  year={2024}
}

@inproceedings{du2024improving,
  title={Improving factuality and reasoning in language models through multiagent debate},
  author={Du, Yilun and Li, Shuang and Torralba, Antonio and Tenenbaum, Joshua B and Mordatch, Igor},
  booktitle={Forty-first international conference on machine learning},
  year={2024}
}

@inproceedings{wang2025mixture,
  title={Mixture-of-agents enhances large language model capabilities},
  author={Wang, Junlin and Wang, Jue and Athiwaratkun, Ben and Zhang, Ce and Zou, James Y},
  booktitle={International Conference on Learning Representations},
  volume={2025},
  pages={33944--33963},
  year={2025}
}

@article{li2024survey,
  title={A survey on LLM-based multi-agent systems: workflow, infrastructure, and challenges},
  author={Li, Xinyi and Wang, Sai and Zeng, Siqi and Wu, Yu and Yang, Yi},
  journal={Vicinagearth},
  volume={1},
  number={1},
  pages={9},
  year={2024},
  publisher={Springer}
}

@article{yang2026towards,
  title={Towards thinking-optimal scaling of test-time compute for llm reasoning},
  author={Yang, Wenkai and Ma, Shuming and Lin, Yankai and Wei, Furu},
  journal={Advances in Neural Information Processing Systems},
  volume={38},
  pages={43605--43631},
  year={2026}
}

@inproceedings{zhao2024expel,
  title={Expel: Llm agents are experiential learners},
  author={Zhao, Andrew and Huang, Daniel and Xu, Quentin and Lin, Matthieu and Liu, Yong-Jin and Huang, Gao},
  booktitle={Proceedings of the AAAI Conference on Artificial Intelligence},
  volume={38},
  number={17},
  pages={19632--19642},
  year={2024}
}

@inproceedings{hong2024metagpt,
  title={MetaGPT: Meta programming for a multi-agent collaborative framework},
  author={Hong, Sirui and Zhuge, Mingchen and Chen, Jonathan and Zheng, Xiawu and Cheng, Yuheng and Wang, Jinlin and Zhang, Ceyao and Yau, Steven and Lin, Zijuan and Zhou, Liyang and others},
  booktitle={International Conference on Learning Representations},
  volume={2024},
  pages={23247--23275},
  year={2024}
}

@article{huang2025survey,
  title={A survey on hallucination in large language models: Principles, taxonomy, challenges, and open questions},
  author={Huang, Lei and Yu, Weijiang and Ma, Weitao and Zhong, Weihong and Feng, Zhangyin and Wang, Haotian and Chen, Qianglong and Peng, Weihua and Feng, Xiaocheng and Qin, Bing and others},
  journal={ACM Transactions on Information Systems},
  volume={43},
  number={2},
  pages={1--55},
  year={2025},
  publisher={ACM New York, NY}
}

@inproceedings{tyen2024llms,
  title={LLMs cannot find reasoning errors, but can correct them given the error location},
  author={Tyen, Gladys and Mansoor, Hassan and C{\u{a}}rbune, Victor and Chen, Yuanzhu Peter and Mak, Tony},
  booktitle={Findings of the Association for Computational Linguistics: ACL 2024},
  pages={13894--13908},
  year={2024}
}

@article{aumann2002incomplete,
  title={Incomplete information},
  author={Aumann, Robert J and Heifetz, Aviad},
  journal={Handbook of game theory with economic applications},
  volume={3},
  pages={1665--1686},
  year={2002},
  publisher={Elsevier}
}

@article{harsanyi1995games,
  title={Games with incomplete information},
  author={Harsanyi, John C},
  journal={American Economic Review},
  volume={85},
  number={3},
  pages={291--303},
  year={1995},
  publisher={Springer}
}

@article{vilares2011bayesian,
  title={Bayesian models: the structure of the world, uncertainty, behavior, and the brain},
  author={Vilares, Iris and Kording, Konrad},
  journal={Annals of the New York Academy of Sciences},
  volume={1224},
  number={1},
  pages={22--39},
  year={2011},
  publisher={Wiley Online Library}
}

@article{Yang2024Qwen25TR,
  title={Qwen2.5 Technical Report},
  author={Qwen An Yang and Baosong Yang and Beichen Zhang and Binyuan Hui and Bo Zheng and Bowen Yu and Chengyuan Li and Dayiheng Liu and Fei Huang and Guanting Dong and Haoran Wei and Huan Lin and Jian Yang and Jianhong Tu and Jianwei Zhang and Jianxin Yang and Jiaxin Yang and Jingren Zhou and Junyang Lin and Kai Dang and Keming Lu and Keqin Bao and Kexin Yang and Le Yu and Mei Li and Mingfeng Xue and Pei Zhang and Qin Zhu and Rui Men and Runji Lin and Tianhao Li and Tingyu Xia and Xingzhang Ren and Xuancheng Ren and Yang Fan and Yang Su and Yi-Chao Zhang and Yunyang Wan and Yuqi Liu and Zeyu Cui and Zhenru Zhang and Zihan Qiu and Shanghaoran Quan and Zekun Wang},
  journal={ArXiv},
  year={2024},
  volume={abs/2412.15115},
  url={https://api.semanticscholar.org/CorpusID:274859421}
}

@misc{mistral_mathstral_2024,
  title        = {Mathstral-7B-v0.1},
  author       = {{Mistral AI}},
  year         = {2024},
  howpublished = {\url{https://huggingface.co/mistralai/Mathstral-7B-v0.1}}
}

@inproceedings{lightman2024let,
  title={Let's verify step by step},
  author={Lightman, Hunter and Kosaraju, Vineet and Burda, Yuri and Edwards, Harrison and Baker, Bowen and Lee, Teddy and Leike, Jan and Schulman, John and Sutskever, Ilya and Cobbe, Karl},
  booktitle={International Conference on Learning Representations},
  volume={2024},
  pages={39578--39601},
  year={2024}
}

@article{chen2021evaluating,
  title={Evaluating large language models trained on code},
  author={Chen, Mark and Tworek, Jerry and Jun, Heewoo and Yuan, Qiming and Pinto, Henrique Ponde De Oliveira and Kaplan, Jared and Edwards, Harri and Burda, Yuri and Joseph, Nicholas and Brockman, Greg and others},
  journal={arXiv preprint arXiv:2107.03374},
  year={2021}
}

@article{wang2024mmlu,
  title={Mmlu-pro: A more robust and challenging multi-task language understanding benchmark},
  author={Wang, Yubo and Ma, Xueguang and Zhang, Ge and Ni, Yuansheng and Chandra, Abhranil and Guo, Shiguang and Ren, Weiming and Arulraj, Aaran and He, Xuan and Jiang, Ziyan and others},
  journal={Advances in Neural Information Processing Systems},
  volume={37},
  pages={95266--95290},
  year={2024}
}

@article{rein2023gpqa,
  title={Gpqa: A graduate-level google-proof q\&a benchmark},
  author={Rein, David and Hou, Betty Li and Stickland, Asa Cooper and Petty, Jackson and Pang, Richard Yuanzhe and Dirani, Julien and Michael, Julian and Bowman, Samuel R},
  journal={arXiv preprint arXiv:2311.12022},
  year={2023}
}

@misc{xie2023pixiu,
 title={PIXIU: A Large Language Model, Instruction Data and Evaluation Benchmark for Finance}, 
 author={Qianqian Xie and Weiguang Han and Xiao Zhang and Yanzhao Lai and Min Peng and Alejandro Lopez-Lira and Jimin Huang},
 year={2023},
 eprint={2306.05443},
 archivePrefix={arXiv},
 primaryClass={cs.CL}
}

@article{wang2022self,
  title={Self-consistency improves chain of thought reasoning in language models},
  author={Wang, Xuezhi and Wei, Jason and Schuurmans, Dale and Le, Quoc and Chi, Ed and Narang, Sharan and Chowdhery, Aakanksha and Zhou, Denny},
  journal={arXiv preprint arXiv:2203.11171},
  year={2022}
}

@article{madaan2023self,
  title={Self-refine: Iterative refinement with self-feedback},
  author={Madaan, Aman and Tandon, Niket and Gupta, Prakhar and Hallinan, Skyler and Gao, Luyu and Wiegreffe, Sarah and Alon, Uri and Dziri, Nouha and Prabhumoye, Shrimai and Yang, Yiming and others},
  journal={Advances in neural information processing systems},
  volume={36},
  pages={46534--46594},
  year={2023}
}

@inproceedings{chen2024reconcile,
  title={Reconcile: Round-table conference improves reasoning via consensus among diverse llms},
  author={Chen, Justin and Saha, Swarnadeep and Bansal, Mohit},
  booktitle={Proceedings of the 62nd Annual Meeting of the Association for Computational Linguistics (Volume 1: Long Papers)},
  pages={7066--7085},
  year={2024}
}

@inproceedings{wang2025agentdropout,
  title={Agentdropout: Dynamic agent elimination for token-efficient and high-performance llm-based multi-agent collaboration},
  author={Wang, Zhexuan and Wang, Yutong and Liu, Xuebo and Ding, Liang and Zhang, Miao and Liu, Jie and Zhang, Min},
  booktitle={Proceedings of the 63rd Annual Meeting of the Association for Computational Linguistics (Volume 1: Long Papers)},
  pages={24013--24035},
  year={2025}
}

@article{yang2025qwen3,
  title={Qwen3 technical report},
  author={Yang, An and Li, Anfeng and Yang, Baosong and Zhang, Beichen and Hui, Binyuan and Zheng, Bo and Yu, Bowen and Gao, Chang and Huang, Chengen and Lv, Chenxu and others},
  journal={arXiv preprint arXiv:2505.09388},
  year={2025}
}

@article{hui2024qwen2,
  title={Qwen2. 5-coder technical report},
  author={Hui, Binyuan and Yang, Jian and Cui, Zeyu and Yang, Jiaxi and Liu, Dayiheng and Zhang, Lei and Liu, Tianyu and Zhang, Jiajun and Yu, Bowen and Lu, Keming and others},
  journal={arXiv preprint arXiv:2409.12186},
  year={2024}
}

@article{jiang20236g,
  title={6G non-terrestrial networks enabled low-altitude economy: Opportunities and challenges},
  author={Jiang, Yihang and Li, Xiaoyang and Zhu, Guangxu and Li, Hang and Deng, Jing and Han, Kaifeng and Shen, Chao and Shi, Qingjiang and Zhang, Rui},
  journal={arXiv preprint arXiv:2311.09047},
  year={2023}
}

@inproceedings{cheng2024instruction,
  title={Instruction pre-training: Language models are supervised multitask learners},
  author={Cheng, Daixuan and Gu, Yuxian and Huang, Shaohan and Bi, Junyu and Huang, Minlie and Wei, Furu},
  booktitle={Proceedings of the 2024 Conference on Empirical Methods in Natural Language Processing},
  pages={2529--2550},
  year={2024}
}

@article{colombo2024saullm,
  title={Saullm-7b: A pioneering large language model for law},
  author={Colombo, Pierre and Pires, Telmo Pessoa and Boudiaf, Malik and Culver, Dominic and Melo, Rui and Corro, Caio and Martins, Andre FT and Esposito, Fabrizio and Raposo, Vera L{\'u}cia and Morgado, Sofia and others},
  journal={arXiv preprint arXiv:2403.03883},
  year={2024}
}

@article{hendrycks2021measuring,
  title={Measuring mathematical problem solving with the math dataset},
  author={Hendrycks, Dan and Burns, Collin and Kadavath, Saurav and Arora, Akul and Basart, Steven and Tang, Eric and Song, Dawn and Steinhardt, Jacob},
  journal={arXiv preprint arXiv:2103.03874},
  year={2021}
}

@inproceedings{chen2021finqa,
  title={Finqa: A dataset of numerical reasoning over financial data},
  author={Chen, Zhiyu and Chen, Wenhu and Smiley, Charese and Shah, Sameena and Borova, Iana and Langdon, Dylan and Moussa, Reema and Beane, Matt and Huang, Ting-Hao and Routledge, Bryan R and others},
  booktitle={Proceedings of the 2021 Conference on Empirical Methods in Natural Language Processing},
  pages={3697--3711},
  year={2021}
}

@article{guha2023legalbench,
  title={Legalbench: A collaboratively built benchmark for measuring legal reasoning in large language models},
  author={Guha, Neel and Nyarko, Julian and Ho, Daniel and R{\'e}, Christopher and Chilton, Adam and Chohlas-Wood, Alex and Peters, Austin and Waldon, Brandon and Rockmore, Daniel and Zambrano, Diego and others},
  journal={Advances in neural information processing systems},
  volume={36},
  pages={44123--44279},
  year={2023}
}

@article{yun2026graph,
  title={Graph-of-agents: A graph-based framework for multi-agent llm collaboration},
  author={Yun, Sukwon and Peng, Jie and Li, Pingzhi and Fan, Wendong and Chen, Jie and Zou, James and Li, Guohao and Chen, Tianlong},
  journal={arXiv preprint arXiv:2604.17148},
  year={2026}
}

@article{li2026hippocampus,
  title={Hippocampus: An efficient and scalable memory module for agentic ai},
  author={Li, Yi and Cao, Lianjie and Ahmed, Faraz and Sharma, Puneet and Li, Bingzhe},
  journal={arXiv preprint arXiv:2602.13594},
  year={2026}
}

@article{xu2026mem,
  title={A-mem: Agentic memory for llm agents},
  author={Xu, Wujiang and Liang, Zujie and Mei, Kai and Gao, Hang and Tan, Juntao and Zhang, Yongfeng},
  journal={Advances in Neural Information Processing Systems},
  volume={38},
  pages={17577--17604},
  year={2026}
}

@article{jiang2026hage,
  title={HAGE: Harnessing Agentic Memory via RL-Driven Weighted Graph Evolution},
  author={Jiang, Dongming and Li, Yi and Li, Guanpeng and Li, Qiannan and Li, Bingzhe},
  journal={arXiv preprint arXiv:2605.09942},
  year={2026}
}

@article{hu2025memory,
  title={Memory in the age of ai agents},
  author={Hu, Yuyang and Liu, Shichun and Yue, Yanwei and Zhang, Guibin and Liu, Boyang and Zhu, Fangyi and Lin, Jiahang and Guo, Honglin and Dou, Shihan and Xi, Zhiheng and others},
  journal={arXiv preprint arXiv:2512.13564},
  year={2025}
}

@article{jiang2026anatomy,
  title={Anatomy of agentic memory: Taxonomy and empirical analysis of evaluation and system limitations},
  author={Jiang, Dongming and Li, Yi and Wei, Songtao and Yang, Jinxin and Kishore, Ayushi and Zhao, Alysa and Kang, Dingyi and Hu, Xu and Chen, Feng and Li, Qiannan and others},
  journal={arXiv preprint arXiv:2602.19320},
  year={2026}
}

@article{jiang2026magma,
  title={MAGMA: A Multi-Graph based Agentic Memory Architecture for AI Agents},
  author={Jiang, Dongming and Li, Yi and Li, Guanpeng and Li, Bingzhe},
  journal={arXiv preprint arXiv:2601.03236},
  year={2026}
}

@article{yang2026groupmembench,
  title={GroupMemBench: Benchmarking LLM Agent Memory in Multi-Party Conversations},
  author={Yang, Jingbo and Lai, Kwei-Herng and Wang, Xiaowen and Chang, Shiyu and Harari, Yaar and Gabrilovich, Evgeniy},
  journal={arXiv preprint arXiv:2605.14498},
  year={2026}
}

@article{yao2022react,
  title={React: Synergizing reasoning and acting in language models},
  author={Yao, Shunyu and Zhao, Jeffrey and Yu, Dian and Du, Nan and Shafran, Izhak and Narasimhan, Karthik and Cao, Yuan},
  journal={arXiv preprint arXiv:2210.03629},
  year={2022}
}

@article{schick2023toolformer,
  title={Toolformer: Language models can teach themselves to use tools},
  author={Schick, Timo and Dwivedi-Yu, Jane and Dess{\`\i}, Roberto and Raileanu, Roberta and Lomeli, Maria and Hambro, Eric and Zettlemoyer, Luke and Cancedda, Nicola and Scialom, Thomas},
  journal={Advances in neural information processing systems},
  volume={36},
  pages={68539--68551},
  year={2023}
}

@article{patil2024gorilla,
  title={Gorilla: Large language model connected with massive apis},
  author={Patil, Shishir G and Zhang, Tianjun and Wang, Xin and Gonzalez, Joseph E},
  journal={Advances in Neural Information Processing Systems},
  volume={37},
  pages={126544--126565},
  year={2024}
}

@article{yao2023tree,
  title={Tree of thoughts: Deliberate problem solving with large language models},
  author={Yao, Shunyu and Yu, Dian and Zhao, Jeffrey and Shafran, Izhak and Griffiths, Tom and Cao, Yuan and Narasimhan, Karthik},
  journal={Advances in neural information processing systems},
  volume={36},
  pages={11809--11822},
  year={2023}
}

@inproceedings{wang2023plan,
  title={Plan-and-solve prompting: Improving zero-shot chain-of-thought reasoning by large language models},
  author={Wang, Lei and Xu, Wanyu and Lan, Yihuai and Hu, Zhiqiang and Lan, Yunshi and Lee, Roy Ka-Wei and Lim, Ee-Peng},
  booktitle={Proceedings of the 61st annual meeting of the association for computational linguistics (volume 1: long papers)},
  pages={2609--2634},
  year={2023}
}

@article{shinn2023reflexion,
  title={Reflexion: Language agents with verbal reinforcement learning},
  author={Shinn, Noah and Cassano, Federico and Gopinath, Ashwin and Narasimhan, Karthik and Yao, Shunyu},
  journal={Advances in neural information processing systems},
  volume={36},
  pages={8634--8652},
  year={2023}
}

@article{shen2023hugginggpt,
  title={Hugginggpt: Solving ai tasks with chatgpt and its friends in hugging face},
  author={Shen, Yongliang and Song, Kaitao and Tan, Xu and Li, Dongsheng and Lu, Weiming and Zhuang, Yueting},
  journal={Advances in Neural Information Processing Systems},
  volume={36},
  pages={38154--38180},
  year={2023}
}

@inproceedings{qin2024toolllm,
  title={Toolllm: Facilitating large language models to master 16000+ real-world apis},
  author={Qin, Yujia and Liang, Shihao and Ye, Yining and Zhu, Kunlun and Yan, Lan and Lu, Yaxi and Lin, Yankai and Cong, Xin and Tang, Xiangru and Qian, Bill and others},
  booktitle={International Conference on Learning Representations},
  volume={2024},
  pages={9695--9717},
  year={2024}
}

@article{lu2023chameleon,
  title={Chameleon: Plug-and-play compositional reasoning with large language models},
  author={Lu, Pan and Peng, Baolin and Cheng, Hao and Galley, Michel and Chang, Kai-Wei and Wu, Ying Nian and Zhu, Song-Chun and Gao, Jianfeng},
  journal={Advances in Neural Information Processing Systems},
  volume={36},
  pages={43447--43478},
  year={2023}
}

@article{wei2022chain,
  title={Chain-of-thought prompting elicits reasoning in large language models},
  author={Wei, Jason and Wang, Xuezhi and Schuurmans, Dale and Bosma, Maarten and Xia, Fei and Chi, Ed and Le, Quoc V and Zhou, Denny and others},
  journal={Advances in neural information processing systems},
  volume={35},
  pages={24824--24837},
  year={2022}
}

@article{wang2023voyager,
  title   = {Voyager: An Open-Ended Embodied Agent with Large Language Models},
  author  = {Guanzhi Wang and Yuqi Xie and Yunfan Jiang and Ajay Mandlekar and Chaowei Xiao and Yuke Zhu and Linxi Fan and Anima Anandkumar},
  year    = {2023},
  journal = {arXiv preprint arXiv: Arxiv-2305.16291}
}

@misc{zhou2024languageagenttreesearch,
      title={Language Agent Tree Search Unifies Reasoning Acting and Planning in Language Models}, 
      author={Andy Zhou and Kai Yan and Michal Shlapentokh-Rothman and Haohan Wang and Yu-Xiong Wang},
      year={2024},
      eprint={2310.04406},
      archivePrefix={arXiv},
      primaryClass={cs.AI},
      url={https://arxiv.org/abs/2310.04406}, 
}

@inproceedings{kwon2023efficient,
  title={Efficient memory management for large language model serving with pagedattention},
  author={Kwon, Woosuk and Li, Zhuohan and Zhuang, Siyuan and Sheng, Ying and Zheng, Lianmin and Yu, Cody Hao and Gonzalez, Joseph and Zhang, Hao and Stoica, Ion},
  booktitle={Proceedings of the 29th symposium on operating systems principles},
  pages={611--626},
  year={2023}
}

@article{hamilton2017inductive,
  title={Inductive representation learning on large graphs},
  author={Hamilton, Will and Ying, Zhitao and Leskovec, Jure},
  journal={Advances in neural information processing systems},
  volume={30},
  year={2017}
}

@article{kipf2016semi,
  title={Semi-supervised classification with graph convolutional networks},
  author={Kipf, Thomas N and Welling, Max},
  journal={arXiv preprint arXiv:1609.02907},
  year={2016}
}

@article{li2026you,
  title={You only spectralize once: Taking a spectral detour to accelerate graph neural network},
  author={Li, Yi and Guo, Zhichun and Li, Guanpeng and Li, Bingzhe},
  journal={Advances in Neural Information Processing Systems},
  volume={38},
  pages={93112--93140},
  year={2026}
}

@inproceedings{li2026veloxgnn,
  title={VeloxGNN: Efficient Out-of-Core GNN Training with Delayed Gradient Propagation},
  author={Li, Yi and Yang, Tsun-Yu and Shen, Zhaoyan and Yang, Ming-Chang and Li, Bingzhe},
  booktitle={2026 IEEE International Symposium on High Performance Computer Architecture (HPCA)},
  pages={1--16},
  year={2026},
  organization={IEEE}
}

\newpage
\appendix
\section{DarkForest as Incomplete-Information Game Theory}\label{app:game_view}
Here we give a formal interpretation of DarkForest through the lens of coordination under incomplete information game theory.

\subsection{Hidden state and private signals}
Let $x$ denote the input and let $Y\in\mathcal{Y}$ denote the unknown correct answer. Before observing any agent output, the system has a prior belief, noted as $p_0(y\mid x)$ over possible answers. Each initial agent $m_i$ produces a private response
$$
r_i \sim m_i(x)
$$
which is parsed into a structured observation
\[
o_i = (a_i,z_i,c_i,v_i,q_i)
\]
Here, $z_i$ is the canonicalized candidate, $c_i$ is the reported or imputed confidence, $v_i$ is the parse-validity indicator, and $q_i$ stores parse-quality metadata. We interpret $o_i$ as a private signal about the hidden state $Y$. The full observation set is
$$
O=\{o_1,\ldots,o_n\}
$$

The key incomplete-information issue is that the system does not observe the latent reliability of each signal. In particular, it does not know whether $z_i=Y$, whether agent $i$ is reliable on the current instance, or whether agreement between two agents reflects independent evidence or correlated error. This is analogous to an incomplete-information game in which agents act with private information and the coordinator must reason over hidden types and signals.

\subsection{Latent agent types}
Let each agent have an unobserved type
$$
\theta_i = (r_i^{\star}, b_i, \kappa_i)
$$
where $r_i^{\star}$ represents the agent's task-dependent reliability, $b_i$ represents possible calibration bias in its confidence, and $\kappa_i$ represents dependence structure with other agents. The coordinator does not observe $\theta_i$ directly. Instead, DarkForest estimates calibration parameters from held-out data:
$$
\Theta =
\left(
\{\alpha_i\}_{i=1}^n,
\{R_{\pi}\}_{\pi\in\Pi},
\lambda_{\mathrm{mal}},
\{\delta_i\}_{i=1}^n
\right)
$$
where $\alpha_i$ estimates agent reliability, $R_{\pi}$ estimates support-pattern reliability, $\lambda_{\mathrm{mal}}$ penalizes malformed but parseable outputs, and $\delta_i$ discounts correlated contributions.

\subsection{Candidate clusters as hypotheses}
After canonicalization, DarkForest forms candidate clusters
$$
C_z=(z,S_z,\pi_z),
\qquad
S_z=\{i:v_i=1,\ z_i=z\}
$$
where $S_z$ is the set of agents supporting candidate $z$ and $\pi_z$ is the support pattern. Each cluster is a hypothesis that $Y=z$. A simple vote would use only $|S_z|$. DarkForest instead assigns a calibrated evidence score
$$
s(z)
=
R_{\pi_z}
\sum_{i\in S_z}
\alpha_i \rho_i \delta_i \phi(c_i)
$$
This can be interpreted as an approximate log-linear evidence model with hand-calibrated sufficient statistics:
$$
\begin{aligned}
&\underbrace{R_{\pi_z}}_{\text{joint pattern reliability}}
\cdot
\underbrace{\sum_{i\in S_z}\alpha_i}_{\text{agent reliability}}\\
&\cdot
\underbrace{\rho_i}_{\text{parse quality}}
\cdot
\underbrace{\delta_i}_{\text{dependence discount}} \cdot
\underbrace{\phi(c_i)}_{\text{confidence modulation}}
\end{aligned}
$$
The support-pattern term acts at the cluster level because it measures the reliability of the joint event that a particular subset of agents supports the same candidate. The dependence correction acts at the individual-contribution level because correlated agents should not be counted as fully independent sources of evidence.

DarkForest normalizes scores into a posterior-like belief distribution:
$$
\hat{p}(Y=z\mid O,x)
=
\frac{\max(0,s(z))}
{\sum_{z'\in\mathcal{C}}\max(0,s(z'))}
$$
We call this distribution posterior-like because it is calibrated from empirical reliability terms rather than derived from a full generative model of agent outputs.

\subsection{Disclosure as an information policy}
A communication-heavy multi-agent method exposes raw responses or reasoning traces across agents. In contrast, DarkForest defines a disclosure operator
$$
D_{\Omega}: (O,B)\mapsto E
$$
where $B$ is the belief state and $\Omega$ is the disclosure policy. The exposed evidence $E$ may include canonical candidates, support patterns, confidence values, posterior mass, margins, and uncertainty indicators, but not full raw traces unless explicitly allowed.

This makes disclosure a controlled information channel. Let
\[
T_{\mathrm{cross}}(E)=\mathrm{tokens}(E)
\]
denote communication cost. A generic decision-theoretic objective for disclosure can be written as
$$
\begin{aligned}
\min_{\Omega}\;
\mathbb{E}_{(x,Y)}
\Big[
&\ell(\hat{Y}_{\Omega}(x),Y) \\
&\quad + \lambda T_{\mathrm{cross}}(D_{\Omega}(O,B)) \\
&\quad + \mu C_{\mathrm{contam}}(\Omega)
\Big]
\end{aligned}
$$
where $\ell$ is task loss, $T_{\mathrm{cross}}$ is token cost, and $C_{\mathrm{contam}}$ is a conceptual penalty for exposing information that can contaminate later decisions. DarkForest does not solve this optimization exactly. Instead, it implements a conservative policy class: preserve independent candidate generation, aggregate signals into a calibrated belief state, and disclose only compact evidence to the coordinator.

\subsection{Coordinator decision and guardrail}
The coordinator receives the input and exposed evidence:
$$
r_{\mathrm{coord}} \sim m_{\mathrm{coord}}(x,E),
\qquad
\hat{z}_{\mathrm{coord}}=\mathcal{P}(r_{\mathrm{coord}})
$$
The coordinator is not treated as an oracle; it is another decision rule operating under incomplete information. DarkForest therefore applies a deterministic guardrail when the belief state strongly supports a conflicting candidate. Let
$$
z^\star=\arg\max_{z\in\mathcal{C}}\hat{p}(Y=z\mid O,x)
$$
and let
$$
\Delta
=
\hat{p}(Y=z^\star\mid O,x)
-
\hat{p}(Y=z^{(2)}\mid O,x)
$$
where $z^{(2)}$ is the runner-up candidate. DarkForest trusts $z^\star$ only when
$$
\begin{aligned}
\mathrm{Trusted}(z^\star)
=
\mathbf{1}\Bigl[
&\ |S_{z^\star}|\ge k \\
&\wedge\; \hat{p}(Y=z^\star\mid O,x)\ge \tau_p \\
&\wedge\; \Delta\ge \tau_m
\Bigr]
\end{aligned}
$$
The final answer is
$$
\hat{Y}
=
\begin{cases}
z^\star,
&
\mathrm{Trusted}(z^\star)=1
\ \wedge\
\hat{z}_{\mathrm{coord}}\ne z^\star\\
\hat{z}_{\mathrm{coord}},
&
\text{otherwise}
\end{cases}
$$
Thus the coordinator proposes a decision from policy-permitted evidence, and the guardrail overrides it only when the calibrated belief state provides strong conflicting evidence.

\subsection{Relation to incomplete-information game theory}
This formulation mirrors three standard ingredients of incomplete-information reasoning: hidden states, private signals, and belief-based decisions. The hidden state is the unknown correct answer $Y$; each agent response is a private signal $o_i$; and DarkForest's belief state is a calibrated summary over candidate hypotheses. The disclosure policy controls which parts of the private-signal summary become public evidence for the coordinator. Unlike a full Bayesian game, DarkForest does not specify utilities for every agent or solve for a Bayesian Nash equilibrium. Instead, it uses the incomplete-information view operationally: preserve private evidence before coordination, discount unreliable or correlated signals, and restrict disclosure when raw communication is costly or may contaminate later decisions.

\section{Detailed Experimental Setup}\label{app:detailed_setup}

\subsection{Backbones}
\label{app:backbone}
We evaluate DarkForest with multiple open-source instruction-tuned language models covering both general-purpose instruction following and domain-specialized tasks. For general-domain tasks, we use Qwen2.5-7B-Instruct~\cite{yang2025qwen3}, Qwen2.5-Coder-7B-Instruct~\cite{hui2024qwen2}, and Mathstral-7B-v0.1~\cite{jiang20236g}. Qwen2.5-7B-Instruct serves as a general instruction-following backbone. Qwen2.5-Coder-7B-Instruct is included to provide stronger code-generation ability, especially for HumanEval. Mathstral-7B-v0.1 is included as a mathematically oriented agent and is used as part of the general-domain agent pool. For professional-domain tasks, we additionally use finance-Llama3-8B~\cite{cheng2024instruction} and Saul-7B-Instruct-v1~\cite{colombo2024saullm}. finance-Llama3-8B is used as a finance-specialized agent, while Saul-7B-Instruct-v1 is used as a legal-domain agent. This mixed-agent setting allows us to evaluate whether DarkForest can coordinate agents with heterogeneous expertise rather than simply aggregating identical model replicas.
\begin{table}[t]
\centering
\small
\caption{Backbone language models used in DarkForest.}
\resizebox{\columnwidth}{!}{%
\begin{tabular}{lll}
\toprule[1.5pt]
Model & Params & Role \\
\midrule[1pt]
Qwen2.5-7B-Instruct        & 7B & general \\
Qwen2.5-Coder-7B-Instruct  & 7B & code \\
Mathstral-7B-v0.1          & 7B & math \\
finance-Llama3-8B          & 8B & finance \\
Saul-7B-Instruct-v1        & 7B & legal \\
\bottomrule[1.5pt]
\end{tabular}%
}
\label{tab:backbones}
\end{table}

\begin{table*}[t]
\centering
\small
\caption{Per-benchmark three-agent composition. All agents in a row are queried independently on every example; DarkForest then constructs the calibrated belief state from their outputs.}
\begin{tabular}{lccccc}
\toprule[1.5pt]
Benchmark & Qwen2.5-7B & Qwen2.5-Coder-7B & Mathstral-7B & finance-Llama3-8B & Saul-7B \\
\midrule[1pt]
MATH        & \checkmark & \checkmark & \checkmark & & \\
HumanEval   & \checkmark & \checkmark & \checkmark & & \\
MMLU-Pro    & \checkmark & \checkmark & \checkmark & & \\
GPQA        & \checkmark & \checkmark & \checkmark & & \\
FinQA       & \checkmark & & & \checkmark & \checkmark \\
LegalBench  & \checkmark & & & \checkmark & \checkmark \\
\bottomrule[1.5pt]
\end{tabular}
\label{tab:agent_composition}
\end{table*}

Table~\ref{tab:backbones} lists the backbones and their HuggingFace identifiers. Table~\ref{tab:agent_composition} gives the exact three-agent composition used on each benchmark.\\
\textbf{Serving and decoding.} All backbones are served via vLLM~\cite{kwon2023efficient} with bf16 weights and tensor-parallel size 1, each model bound to a single H100 GPU. Servers expose OpenAI-compatible local chat endpoints. Unless otherwise specified, we use deterministic or low-temperature decoding: candidate generation uses temperature $\tau \in \{0.0, 0.2, 0.7\}$ depending on the benchmark (specified per-benchmark in the run scripts), top-$p = 1.0$, and \texttt{max\_tokens} in $\{800, 1024, 2048\}$ matched to the expected answer length. The base random seed is 0, with per-agent offsets ($+100{,}000$, $+200{,}000$) to make stochastic decoding reproducible across parallel candidates. The underlying model decoding settings are held fixed across DarkForest and baseline methods on the same benchmark whenever the method permits.

\subsection{Benchmarks}
\label{app:benchmark}
\begin{table*}[t]
\centering
\small
\caption{Benchmark splits, calibration sizes (for DarkForest's calibration stage), evaluation sizes, and primary metrics. GoA-aligned denotes the same splits released by the Graph-of-Agents codebase, which we adopt for direct comparability}
\begin{tabular}{llrrl}
\toprule[1.5pt]
Benchmark & Split & Calibration & \# Eval & Metric \\
\midrule[1pt]
MATH        & GoA-aligned test subset~\cite{yun2026graph}     & 100 & 500     & Exact match \\
HumanEval   & GoA-aligned dev subset~\cite{yun2026graph}      & 114 & 50      & Pass@1 \\
MMLU-Pro    & GoA-aligned sampled test~\cite{yun2026graph}    & 70 & 2{,}100 & Accuracy \\
GPQA        & GoA-aligned~\cite{yun2026graph} & 50 & 198 & Accuracy \\
FinQA       & text-only support subset of FinQA test          & 100                & 300     & Execution / Program accuracy \\
LegalBench  & 154 exact-match-eligible tasks                  & 100                & 500     & Exact match \\
\bottomrule[1.5pt]
\end{tabular}
\label{tab:benchmarks}
\end{table*}

We evaluate DarkForest on six benchmarks that spanning mathematical reasoning, code generation, general scientific knowledge reasoning, graduate-level reasoning, financial question answering, and legal reasoning. The benchmark suite is designed to test whether controlled communication is useful across both general-domain and professional-domain multi-agent reasoning. Table~\ref{tab:benchmarks} summarizes the splits, sample counts, and metrics.\\
\textbf{MATH} evaluates mathematical problem solving over competition-style questions~\cite{hendrycks2021measuring}. We evaluate on a 500-sample subset aligned with the Graph-of-Agents (GoA) release~\cite{yun2026graph}. We use exact match after answer normalization as the default metric.\\
\textbf{HumanEval} evaluates code generation ability~\cite{chen2021evaluating}. We report Pass@1 (single sample, $n=1$) on the 50-problem dev subset aligned with the GoA release~\cite{yun2026graph}; generated programs are executed against the reference unit tests bundled with HumanEval via the standard \texttt{human-eval} evaluation harness.\\
\textbf{MMLU-Pro} evaluates challenging multiple-choice reasoning across broad academic and professional subjects~\cite{wang2024mmlu}. We evaluate on the 2{,}100-question sampled test split released with GoA~\cite{yun2026graph} and report accuracy. Compared with simpler multiple-choice datasets, MMLU-Pro requires stronger reasoning and is therefore suitable for evaluating whether DarkForest can improve final answer selection under disagreement among agents.\\
\textbf{GPQA} evaluates difficult graduate-level scientific question answering~\cite{rein2023gpqa}. We adopt the GPQA split released by Graph-of-Agents (GoA)~\cite{yun2026graph}, comprising 198 evaluation questions and report accuracy. GPQA is included because it is a high-uncertainty benchmark where naive majority voting may be unreliable when agents share similar misconceptions; this setting is well aligned with DarkForest's motivation of treating agreement as calibrated evidence rather than as an unweighted vote.\\
\textbf{FinQA} evaluates financial question answering requiring numerical reasoning over financial reports~\cite{chen2021finqa}. To isolate the language-reasoning component from table-parsing artifacts and to keep prompts within context length, we restrict evaluation to the text-only support subset of the FinQA test split. This yields 300 evaluation examples and a disjoint pool of 100 calibration examples. We report both execution accuracy and program accuracy: execution accuracy measures whether the final computed answer is correct, while program accuracy measures whether the generated reasoning program matches the expected computation.\\
\textbf{LegalBench} evaluates legal reasoning across tasks derived from legal texts and legal decision-making settings~\cite{guha2023legalbench}. We restrict to the 154 LegalBench tasks whose evaluation protocol is single-label exact match, excluding 8 generation or free-form tasks, and stratified-sample 500 evaluation rows and 100 calibration rows across these tasks. We report micro-averaged exact match over the sampled rows. The label space is heterogeneous: the majority of tasks are binary (yes/no), while a minority are small multi-class tasks, such as \texttt{abercrombie} with labels \{arbitrary, descriptive, fanciful, generic, suggestive\}.

\subsection{Baselines}
\label{app:baselines}
We compare our DarkForest against the following multi-agent baselines, all run with the same benchmark-specific backbone pool and model decoding settings as DarkForest whenever applicable.

\begin{itemize}[topsep=0pt,itemsep=-1ex,partopsep=1ex,parsep=1ex, leftmargin=*]
\item \textbf{Self-consistency (majority vote).} Majority vote over the three independent agent outputs that DarkForest itself uses, with uniform random tie-breaking under seed 0~\cite{wang2022self}. This isolates the contribution of DarkForest's calibrated aggregation over a simple vote.
\item \textbf{Graph-of-Agents (GoA).} Graph-structured multi-agent collaboration~\cite{yun2026graph}; we use the official splits and protocols.
\item \textbf{Mixture-of-Agents (MoA).} Layered aggregation of agent outputs~\cite{wang2025mixture}.
\item \textbf{ReConcile.} Multi-round agent discussion with confidence-weighted voting~\cite{chen2024reconcile}.
\item \textbf{LLM Multi-Agent Debate.} Iterative debate among agents with answer revision across rounds~\cite{du2024improving}.
\item \textbf{Self-Refine.} Single-agent iterative self-critique and revision~\cite{madaan2023self}.
\end{itemize}
All methods use the same task-level prompt and answer-extraction format when possible. Method-specific coordination prompts follow the corresponding baseline implementation when available.

\section{Detailed Related Work}\label{app: detailed_related}

LLM agent research spans a broad set of mechanisms for extending language models beyond single-turn generation, including tool use, planning, memory, role specialization, and inter-agent communication. Early tool-augmented and action-oriented agents connect language models to external APIs, search engines, calculators, code interpreters, or symbolic tools, allowing models to act in an environment rather than only produce text~\cite{yao2022react,schick2023toolformer,patil2024gorilla,shen2023hugginggpt,qin2024toolllm,lu2023chameleon}. Planning-oriented methods further decompose complex goals into intermediate steps or search over possible reasoning trajectories before committing to an answer~\cite{wei2022chain,yao2023tree,wang2023plan,shinn2023reflexion,wang2023voyager,zhou2024languageagenttreesearch}. These works show that agent performance often depends not only on the underlying model, but also on how information is represented, routed, and exposed during decision-making.

Another related line of work studies memory-augmented agents, where the main challenge is how to organize long interaction histories into structured and retrievable evidence rather than exposing unfiltered context to the model~\cite{xu2026mem,li2026hippocampus,jiang2026hage,hu2025memory,jiang2026magma,yang2026groupmembench, jiang2026anatomy}. General multi-agent frameworks such as CAMEL, AutoGen, and MetaGPT instantiate agents as role-conditioned or workflow-driven participants that exchange messages to solve complex tasks~\cite{li2023camel,wu2024autogen,hong2024metagpt}. DarkForest addresses a different but connected problem: given several independently generated agent outputs for the same test-time question, what evidence should be disclosed to the coordinator, and how should that evidence be calibrated before disclosure?

We therefore focus our detailed comparison on multi-agent coordination baselines that appear in Table~\ref{tab:overall_comparison}. These methods differ primarily in how information crosses agent boundaries: interaction-based methods exchange reasoning between agents, aggregation-based methods combine independently produced candidates, and refinement or efficiency-oriented methods modify a single trajectory or prune redundant agents.\\
\textbf{LLM Multi-Agent Debate~\cite{du2024improving}.} Multiple LLM agents independently propose answers to a question, then engage in several rounds of debate in which each agent reads the other agents' reasoning and revises its own response. The final answer is taken from the agent outputs after the debate converges. Because each round exposes full reasoning traces, an early-round error or a persuasive but incorrect argument can be adopted and amplified by other agents; debate also incurs substantial token overhead proportional to the number of agents and rounds. DarkForest removes the inter-round exchange entirely: agents do not observe one another's outputs, and disagreement is resolved through a calibrated belief state rather than through textual persuasion.\\
\textbf{ReConcile~\cite{chen2024reconcile}.}
ReConcile organizes multiple LLM agents in a round-table discussion in which each agent reports an answer with a self-reported confidence value, then revises its answer after seeing the others' responses; the final answer is selected by confidence-weighted voting. This relaxes pure debate by introducing explicit confidence, but the confidence values are not calibrated against ground truth, and the inter-agent exchange still allows error propagation. DarkForest instead calibrates per-agent reliability and per-support-pattern reliability on a held-out set, treats raw confidence as one input to an evidence score rather than as the deciding factor, and keeps candidate generation independent.\\
\textbf{Self-Consistency~\cite{wang2022self}.}
Self-Consistency samples multiple chain-of-thought reasoning traces from a single model under stochastic decoding and returns the majority-voted final answer. It is widely used because it is simple and avoids cross-agent communication, but it treats all samples as exchangeable and weighs every agreement equally regardless of which models agreed and how reliable they have historically been. DarkForest generalizes this aggregation step: it allows heterogeneous agents, weights each agent's support by its calibrated reliability, weights agreement by support-pattern reliability, and applies independence corrections to discount correlated agreement.\\
\textbf{Mixture-of-Agents (MoA)~\cite{wang2025mixture}.}
MoA arranges multiple LLM agents in a layered architecture: a first layer produces candidate responses to the input, and one or more subsequent aggregator layers read these candidates and synthesize an improved response. This shifts aggregation into the language space, which can produce fluent combined answers, but the aggregator receives raw candidate text and may inherit hallucinations or confidently incorrect content from any single proposer. DarkForest replaces text-level synthesis with a structured belief interface: only parsed canonical candidates, support patterns, posterior mass, and uncertainty indicators are exposed to the coordinator, which treats them as evidence rather than as text to be merged.\\
\textbf{Graph-of-Agents (GoA)~\cite{yun2026graph}.} Inspired from Graph Neural Network (GNN)~\cite{hamilton2017inductive,kipf2016semi,li2026you,li2026veloxgnn},
GoA places agents on a graph and routes candidate outputs and intermediate messages along graph edges, with aggregation operators (mean and max, in the two variants we evaluate) combining the per-node outputs. GoA captures structured dependencies among agents but, like MoA, performs aggregation directly over candidate outputs without an explicit reliability model, and its mean and max operators do not distinguish between high-confidence agreement and accidental agreement. DarkForest is complementary in spirit but operates on a calibrated posterior over candidate clusters; the calibrated scores can be interpreted as a learned, support-pattern-aware aggregation that subsumes both mean and max as special cases.\\
\textbf{Self-Refine~\cite{madaan2023self}.}
Self-Refine is a single-agent baseline in which one LLM iteratively produces an output, critiques it, and rewrites it until a stopping criterion is met. It targets a different axis from the multi-agent methods above: it improves a single trajectory through self-feedback rather than aggregating across independent agents. We include it as a strong single-agent reference. DarkForest differs in kind: it does not refine an output in place but instead aggregates evidence from multiple independent agents under a controlled disclosure policy.

\section{Ablation Study}\label{app:ablation}
This appendix provides component-level ablations for DarkForest. Here we isolate individual design choices in the final decision procedure.

\subsection{Voting ablation}\label{sec:vote}
\begin{table}[!t]
\centering
\caption{Voting ablation on two benchmarks: MMLU-Pro and LegalBench.}
\label{tab:ablation_voting}
\resizebox{\linewidth}{!}{%
\begin{tabular}{lccc}
\toprule[1.3pt]
\multirow{2}{*}{Dataset}
& \multicolumn{3}{c}{Quality} \\
\cmidrule(lr){2-4}
& Majority Vote & Weighted Vote & Full DarkForest \\
\midrule[0.8pt]
MMLU-Pro   & 53.00 & 56.38 & 58.38 \\
LegalBench & 65.00 & 66.60 & 68.00 \\
\bottomrule[1.3pt]
\end{tabular}%
}
\end{table}
\begin{figure}[t]
\centering
\includegraphics[width=\columnwidth]{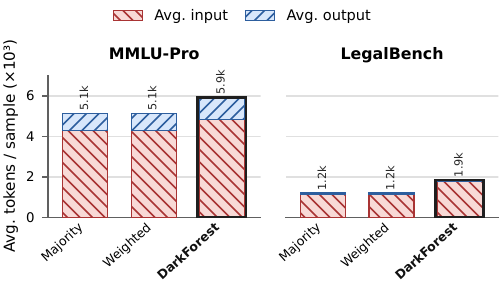}
\caption{Token consumption for voting ablation.}
\label{fig:ablation_voting_tokens}
\vspace{-12pt}
\end{figure}
We first test whether DarkForest reduces to a voting rule. Table~\ref{tab:ablation_voting} compares Full DarkForest with two voting baselines on MMLU-Pro and LegalBench. Majority Vote selects the largest candidate cluster with uniform agent weights, while Weighted Vote uses calibrated agent reliability but does not use the full belief construction, coordinator, or guardrail. DarkForest consistently improves over both voting baselines. On MMLU-Pro, Majority Vote obtains 53.00\% accuracy, Weighted Vote improves to 56.38\%, and Full DarkForest further improves to 58.38\%. On LegalBench, the same trend holds: Majority Vote reaches 65.00\%, Weighted Vote reaches 66.60\%, and Full DarkForest reaches 68.00\%. These results show that reliability calibration is useful, but not sufficient by itself. The additional gains of DarkForest come from treating agreement as structured evidence, incorporating support-pattern and uncertainty signals, and allowing the coordinator to verify the calibrated belief state rather than directly returning the most frequent answer. Figure~\ref{fig:ablation_voting_tokens} shows that this improvement comes with only a modest token increase. Majority Vote and Weighted Vote use the same independent agent outputs and therefore have identical token cost. DarkForest adds a compact coordinator call, increasing token usage from 5.1k to 5.9k on MMLU-Pro and from 1.2k to 1.9k on LegalBench. The quality gains therefore do not rely on unrestricted inter-agent communication or multi-round discussion; they come from a small amount of controlled disclosure.

\subsection{Coordinator and Guardrail}\label{app:coordinator_guardrail}
\begin{table}[t]
\centering
\caption{Decision mechanism ablation. w/o Coordinator directly returns the top-belief candidate without a coordinator call, and w/o Guardrail keeps the coordinator but removes the deterministic post-processing step.}
\label{tab:ablation_decision}
\resizebox{\linewidth}{!}{%
\begin{tabular}{lccc}
\toprule[1.5pt]
\multirow{2}{*}{Dataset}
& \multicolumn{3}{c}{Quality} \\
\cmidrule(lr){2-4}
& \begin{tabular}[c]{@{}c@{}}DarkForest\\w/o Coordinator\end{tabular}
& \begin{tabular}[c]{@{}c@{}}DarkForest\\w/o Guardrail\end{tabular}
& \begin{tabular}[c]{@{}c@{}}Full\\DarkForest\end{tabular} \\
\midrule[1pt]
MATH       & 76.80 & 75.40 & 76.80 \\
LegalBench & 67.20 & 65.60 & 68.00 \\
\bottomrule[1.5pt]
\end{tabular}%
}
\end{table}
\begin{figure}[t]
\centering
\includegraphics[width=\columnwidth]{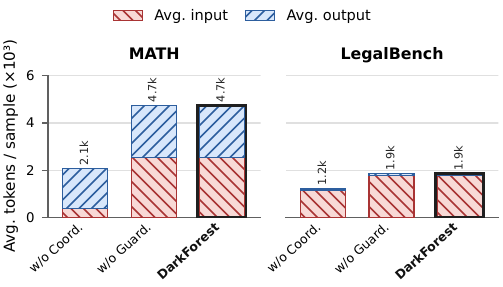}
\caption{Average token consumption for the decision mechanism ablation.}
\label{fig:ablation_decision_tokens}
\end{figure}
We now isolate the two final-decision components of DarkForest: the coordinator and the deterministic guardrail. Table~\ref{tab:ablation_decision} compares Full DarkForest with two variants. The first removes the coordinator and directly returns the top-belief candidate. The second keeps the coordinator but removes the deterministic guardrail. The results show that the coordinator and guardrail play complementary roles. Removing the coordinator has no effect on MATH, where the calibrated belief state is already sufficiently decisive, but reduces LegalBench accuracy from 68.00\% to 67.20\%. This suggests that the coordinator is most useful when candidate labels are heterogeneous or when the task benefits from checking compact evidence against the original input. In contrast, removing the guardrail hurts both benchmarks: MATH drops from 76.80\% to 75.40\%, and LegalBench drops from 68.00\% to 65.60\%. The guardrail therefore protects the system from cases where the coordinator output conflicts with a strongly supported belief cluster. Figure~\ref{fig:ablation_decision_tokens} reports the corresponding token cost. The variant without a coordinator is cheaper because it avoids the final coordinator call, using 2.1k tokens on MATH and 1.2k on LegalBench. Full DarkForest uses the same token budget as the no-guardrail variant because the guardrail is deterministic post-processing and introduces no model call. This confirms that the guardrail improves quality without increasing token consumption.

\subsection{Disclosure}
\begin{table}[t]
\centering
\caption{Disclosure policy ablation on GPQA.}
\label{tab:ablation_disclosure_gpqa}
\resizebox{\linewidth}{!}{%
\begin{tabular}{lcccc}
\toprule[1.3pt]
Disclosure Policy & Quality & Avg Input & Avg Output & Avg Total \\
\midrule[0.8pt]
Belief summary & 40.00 & 1368.7 & 2066.7 & 3435.5 \\
Reasoning summary & 40.00 & 2084.2 & 2051.8 & 4136.0 \\
Full raw traces & 36.67 & 2920.9 & 2083.7 & 5004.5 \\
\bottomrule[1.3pt]
\end{tabular}%
}
\end{table}
We study how much information should be exposed to the coordinator. Table~\ref{tab:ablation_disclosure_gpqa} compares three disclosure policies on GPQA dataset. Belief summary (our current design) exposes only compact structured evidence, including canonical candidates, support patterns, posterior mass, margins, and uncertainty indicators. Reasoning summary additionally exposes short natural-language summaries of agent rationales. Full raw traces exposes the complete initial agent responses to the coordinator.

The results show that exposing more text does not improve final quality. Belief summary and reasoning summary both achieve 40.00\% accuracy, but belief summary uses fewer tokens: 3435.5 total tokens per sample compared with 4136.0 for reasoning summary. Full raw traces further increase the total cost to 5004.5 tokens per sample, but reduce accuracy to 36.67\%. This supports the main design choice of DarkForest: the coordinator benefits from compact calibrated evidence, while exposing longer raw traces can increase cost and may introduce distracting or misleading intermediate reasoning.

\subsection{Calibration}
\begin{figure}[t]
\centering
\includegraphics[width=\columnwidth]{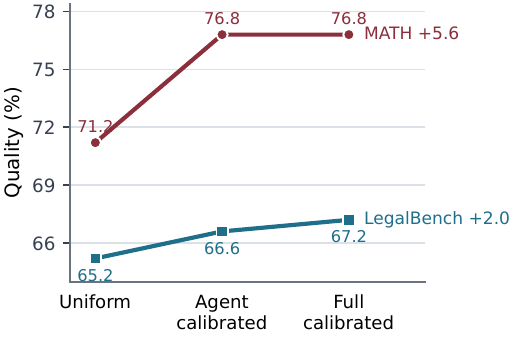}
\caption{Calibration ablation on MATH and LegalBench. Uniform uses equal agent and support-pattern weights.
Agent calibrated uses calibrated per-agent reliability.}
\label{fig:calibration_ablation}
\end{figure}
We ablate calibration components in the belief construction. Figure~\ref{fig:calibration_ablation} compares three variants. Uniform assigns equal weight to all valid agent supports. Agent calibrated uses calibrated per-agent reliability but does not use the full set of belief-state signals, while the Full calibrated uses the complete DarkForest belief construction, including agent reliability, support-pattern reliability, parse-quality penalties, confidence weighting, and independence correction.

From Figure~\ref{fig:calibration_ablation}, we can see that calibration consistently improves quality. On MATH, uniform weighting achieves 71.2\%, while agent-level calibration and full calibration both reach 76.8\%, giving a 5.6-point gain. On LegalBench, uniform weighting reaches 65.2\%, agent-level calibration improves to 66.6\%, and full calibration further improves to 67.2\%. These results show that calibration is useful beyond simple support counting. Agent reliability is the dominant factor on MATH, where answer canonicalization is relatively clean, while the full belief construction provides additional benefit on LegalBench, where task labels and agent behavior are more heterogeneous.

\subsection{Guardrail Threshold Sensitivity}\label{sec:threshold_sen}
\begin{table}[t]
\centering
\caption{Guardrail threshold sensitivity. We vary the posterior threshold $\tau_p$ while keeping the margin threshold $\tau_m$ fixed at the default value for each benchmark. Since guardrail decisions are deterministic post-processing, all rows for the same benchmark have the same LLM token consumption as Full DarkForest. Quality, override rate, and wrong override rate are reported in percentages.}
\label{tab:ablation_guardrail_sensitivity}
\resizebox{\columnwidth}{!}{%
\begin{tabular}{lccccc}
\toprule[1.5pt]
Dataset & $\tau_p$ & $\tau_m$ & Quality & Override Rate & Wrong Override / Sample \\
\midrule[1pt]
MATH       & 0.50 & 0.25 & 76.80 & 13.80 & 3.20 \\
MATH       & 0.66 & 0.25 & 76.80 & 13.80 & 3.20 \\
MATH       & 0.80 & 0.25 & 76.20 & 13.00 & 3.00 \\
\midrule
LegalBench & 0.50 & 0.20 & 68.00 & 7.60 & 2.40 \\
LegalBench & 0.66 & 0.20 & 68.00 & 7.60 & 2.40 \\
LegalBench & 0.80 & 0.20 & 66.60 & 1.80 & 0.40 \\
\bottomrule[1.5pt]
\end{tabular}%
}
\end{table}
We test whether the guardrail depends on a finely tuned posterior threshold. Table~\ref{tab:ablation_guardrail_sensitivity} varies the posterior threshold $\tau_p$ while keeping the margin threshold $\tau_m$ fixed at its default value for each benchmark. Since the guardrail is deterministic post-processing, all rows for the same benchmark have the same LLM token consumption. The results show that the default threshold is stable. On MATH, $\tau_p=0.50$ and $\tau_p=0.66$ both achieve 76.80\% accuracy, with the same override rate of 13.80\% and wrong-override rate of 3.20\%. Increasing the threshold to 0.80 lowers the override rate slightly, but also reduces accuracy to 76.20\%, suggesting that the guardrail becomes too conservative and fails to correct some coordinator mistakes. LegalBench shows the same pattern: $\tau_p=0.50$ and $\tau_p=0.66$ both achieve 68.00\%, while $\tau_p=0.80$ reduces accuracy to 66.60\%. The higher threshold sharply reduces the override rate from 7.60\% to 1.80\%, but this reduction removes useful corrections along with wrong overrides. These results support the use of a moderate posterior threshold. The guardrail should not fire on weak evidence, but it should also not require near-unanimous confidence. A threshold around two thirds provides a conservative strong-evidence criterion: it preserves the quality gains of the guardrail while keeping wrong overrides limited.

\subsection{Scalability}
\begin{table}[t]
\centering
\caption{Scalability ablation on FinQA. DarkForest scales the independent agent pool from three to five backbones.}
\label{tab:group7_finqa_scalability_quality}
\resizebox{\columnwidth}{!}{%
\begin{tabular}{lcc}
\toprule[1.5pt]
Setting & Execution Accuracy (\%) & Program Accuracy (\%) \\
\midrule[1pt]
3 backbones (default) & 15.67 & 11.33 \\
4 backbones (+Qwen-Coder) & 19.00 & 13.00 \\
5 backbones (+Qwen-Coder, +Mathstral) & 19.33 & 13.67 \\
\bottomrule[1.5pt]
\end{tabular}%
}
\end{table}
\begin{figure}[t]
\centering
\includegraphics[width=\columnwidth]{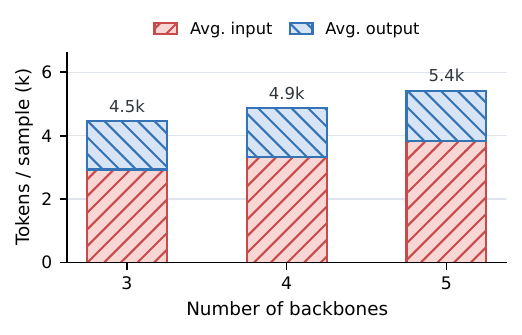}
\caption{Average token consumption in the FinQA scalability ablation.}
\label{fig:scalable_ablation}
\end{figure}
We study whether DarkForest can benefit from a larger pool of independent backbones. The default FinQA setting uses three agents (See Section~\ref{sec:overall_comp}). We then add Qwen2.5-Coder-7B-Instruct (Qwen-Coder) as a fourth backbone and Mathstral-7B-v0.1 (Mathstral) as a fifth backbone, while keeping the same controlled-disclosure decision procedure. This ablation tests whether DarkForest can incorporate additional candidate evidence without turning coordination into unrestricted multi-round communication.

Table~\ref{tab:group7_finqa_scalability_quality} shows that adding more independent backbones improves both FinQA metrics. Moving from three to four backbones increases execution accuracy from 15.67\% to 19.00\% and program accuracy from 11.33\% to 13.00\%. Adding a fifth backbone gives a smaller but still positive gain, reaching 19.33\% execution accuracy and 13.67\% program accuracy. Figure~\ref{fig:scalable_ablation} shows that the corresponding token increase is moderate: average total token usage grows from 4.5k to 4.9k and 5.4k tokens per sample. These results indicate that DarkForest can use additional independent evidence when available, while keeping communication compact through the same belief-summary interface.

\subsection{Coordinator Robustness}
\begin{figure}[t]
\centering
\includegraphics[width=\columnwidth]{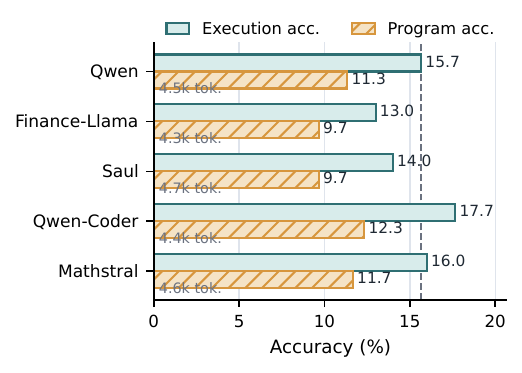}
\caption{Coordinator robustness ablation on FinQA. We fix the initial agent traces, calibrated belief state, and deterministic guardrail, and vary only the coordinator model.
Bars report execution accuracy and program accuracy; text labels report average total token consumption per sample.}
\label{fig:coordinator_robustness}
\end{figure}
We evaluate whether DarkForest depends on a specific coordinator model. In this ablation, we use the full FinQA evaluation set and fix the initial agent traces, calibrated belief state, disclosure policy, and deterministic guardrail. We then replace only the coordinator model, using Qwen2.5-7B-Instruct (Qwen), finance-Llama3-8B (Finance-Llama), Saul-7B-Instruct-v1 (Saul), Qwen2.5-Coder-7B-Instruct (Qwen-Coder), and Mathstral-7B-v0.1 (Mathstral) as alternatives. Since FinQA evaluates both final numerical correctness and generated reasoning programs, we report execution accuracy and program accuracy.

Figure~\ref{fig:coordinator_robustness} shows that the default Qwen coordinator is not the only effective choice. Qwen-Coder improves over the default, reaching 17.67\% execution accuracy and 12.33\% program accuracy, while Mathstral also remains competitive with 16.00\% execution accuracy and 11.67\% program accuracy. Finance-Llama and Saul perform worse, suggesting that coordinator choice still matters. Overall, these results show that DarkForest does not rely on a single fixed coordinator model: several coordinators can use the same calibrated belief state effectively, while weaker coordinators degrade quality but do not change the underlying controlled-disclosure procedure.

\end{document}